\documentclass{sig-alternate}

\pdfoutput=1
\usepackage{subcaption}
\usepackage{standalone}
\usepackage{color}
\usepackage{color, colortbl}
\usepackage{multirow}
\definecolor{high}{rgb}{0.7,.85,.85}
\usepackage[table]{xcolor}
\usepackage{siunitx}
\definecolor{lightgray}{gray}{0.9}
\usepackage{microtype}
\usepackage{pifont}

\usepackage{framed}
\usepackage{mathtools,xparse}
\usepackage{tikz}
\usepackage{pgfplots}
\usepackage[hidelinks]{hyperref}
\usepackage{booktabs}
\usepackage{textcomp}
\usepackage[numbers,sort&compress]{natbib}
\usepackage{algorithmicx}
\usepackage{algorithm}
\usepackage{algcompatible}

\usepackage[skip=0pt]{caption}

\DeclarePairedDelimiter{\norm}{\lVert}{\rVert}
\NewDocumentCommand{\normL}{ s O{} m }{%
  \IfBooleanTF{#1}{\norm*{#3}}{\norm[#2]{#3}}_{L_2(\Omega)}%
}

\definecolor{ourgreen}{rgb}{0.0,0.5,0.0}
\newcommand{\comment}[1]{}

\newcommand{\freq}{\emph{Frequency}}
\newcommand{\syn}{\emph{Syntactic}}
\newcommand{\ngrams}{Google Books Ngram Corpus}
\newcommand{\dist}{\emph{Distributional}}
\newcommand{\hwplotA}{\raisebox{2pt}{\tikz{\draw[red,solid,line width=2pt](0,0) -- (5mm,0);}}}
\newcommand{\hwplotB}{\raisebox{2pt}{\tikz{\draw[ourgreen,solid,line width=2pt](0,0) -- (5mm,0);}}}
\newcommand{\hwplotC}{\raisebox{2pt}{\tikz{\draw[blue,solid,line width=2pt](0,0) -- (5mm,0);}}}

\newcommand{\pvalue}{$p$-value}

\usepackage{amsmath}
\DeclareMathOperator*{\argmin}{argmin}

\begin{document}
\title{Statistically Significant Detection of Linguistic Change }
%
%
%
%
%

\numberofauthors{2} 
%
\author{
%
%
\alignauthor
Vivek Kulkarni\\
       \affaddr{Stony Brook University, USA}\\
       \email{vvkulkarni@cs.stonybrook.edu}
\alignauthor
Rami Al-Rfou\\
       \affaddr{Stony Brook University, USA}\\
       \email{ralrfou@cs.stonybrook.edu}
\and
\alignauthor
Bryan Perozzi\\
       \affaddr{Stony Brook University, USA}\\
       \email{bperozzi@cs.stonybrook.edu}
\alignauthor
Steven Skiena\\
       \affaddr{Stony Brook University, USA}\\
       \email{skiena@cs.stonybrook.edu}
}

\maketitle

\begin{abstract}
We propose a new computational approach for tracking and detecting statistically significant linguistic shifts in the meaning and usage of words. Such linguistic shifts are especially prevalent on the Internet, where the rapid exchange of ideas can quickly change a word's meaning. Our meta-analysis approach constructs property time series of word usage, and then uses statistically sound change point detection algorithms to identify significant linguistic shifts.

We consider and analyze three approaches of increasing complexity to generate such linguistic property time series, the culmination of which uses distributional
characteristics inferred from word co-occurrences. Using recently proposed deep neural language models, we first train vector representations of words for each time period. Second, we warp the vector spaces into one unified coordinate system. Finally, we construct a distance-based distributional time series for each word to track it's linguistic displacement over time.

We demonstrate that our approach is scalable by tracking linguistic change across years of micro-blogging using Twitter, a decade of product reviews using a corpus of movie reviews from Amazon, and a century of written books using the Google Book-ngrams. Our analysis reveals interesting patterns of language usage change commensurate with each medium.
\end{abstract}



\vfill

\section{Introduction}
\label{sec:Introduction}

Natural languages are inherently dynamic, evolving over time to accommodate the needs of their speakers.
This effect is especially prevalent on the Internet, where the rapid exchange of ideas can change a word's meaning overnight.

In this paper, we study the problem of detecting such linguistic shifts on a variety of media including micro-blog posts, product reviews, and books.
Specifically, we seek to detect the broadening and narrowing of semantic senses of words, as they  continually change throughout the lifetime of a medium. 

\begin{figure}[t!]
  \includegraphics[width=\columnwidth]{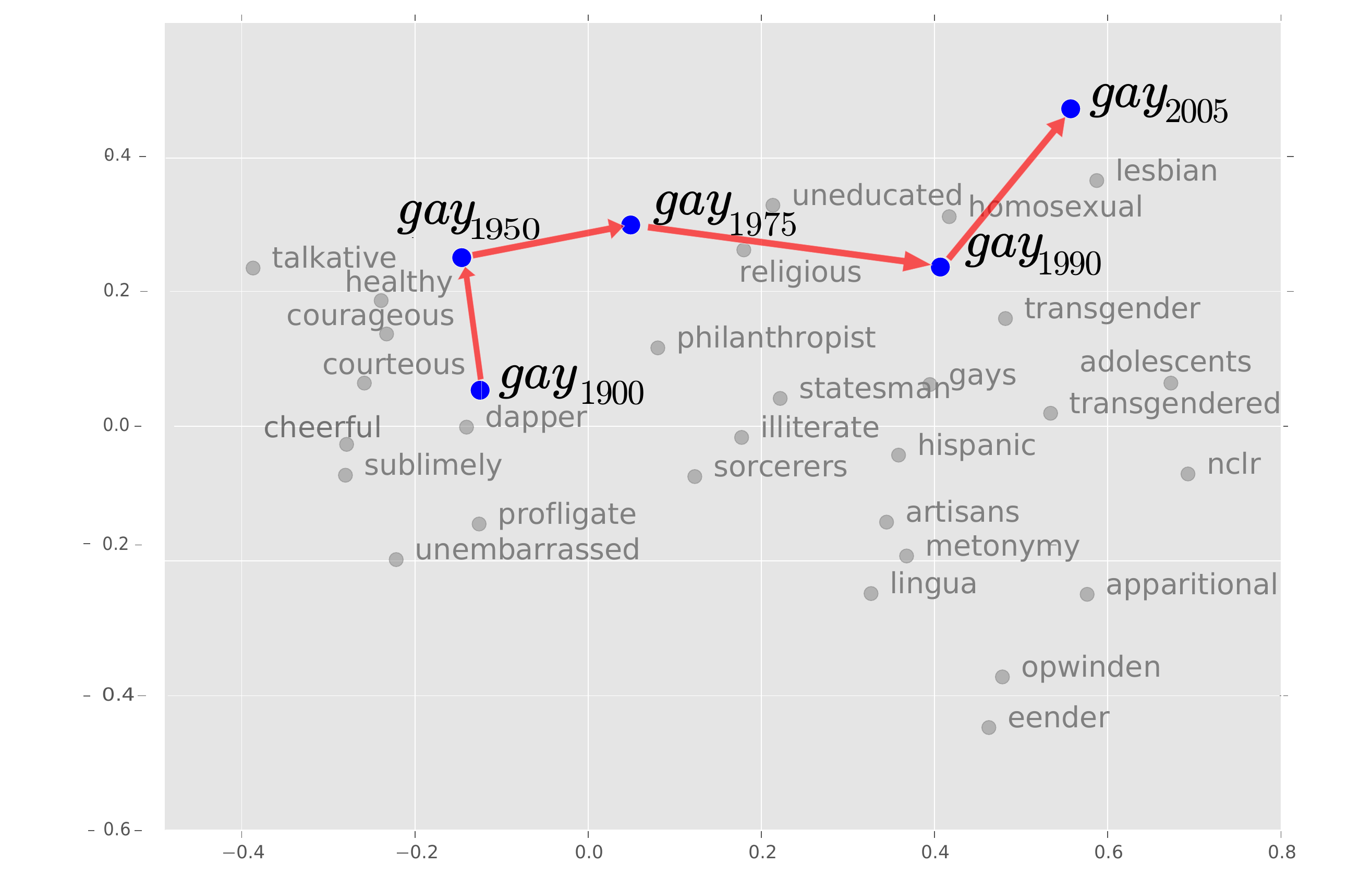}
  \vspace{0.01in}
  \caption{A 2-dimensional projection of the latent semantic space captured by our algorithm.  Notice the semantic trajectory of the word \texttt{gay} transitioning meaning in the space.}
  \label{fig:vis_embeddings}
\end{figure}

We propose the first computational approach for tracking and detecting statistically significant linguistic shifts of words.
To model the temporal evolution of natural language, we construct a time series per word.
We investigate three methods to build our word time series.
First, we extract \freq\ based statistics to capture sudden changes in word usage.
Second, we construct \syn\ time series by analyzing each word's part of speech (POS) tag distribution.
Finally, we infer contextual cues from word co-occurrence statistics to construct \dist\ time series.
In order to detect and establish statistical significance of word changes over time, we present a change point detection algorithm, which is compatible with all methods.

Figure \ref{fig:vis_embeddings} illustrates a 2-dimensional projection of the latent semantic space captured by our \dist\ method.
We clearly observe the sequence of semantic shifts that the word \texttt{gay} has undergone over the last century (1900-2005).
Initially, \texttt{gay} was an adjective that meant \texttt{cheerful} or \texttt{dapper}. Observe for the first 50 years, that it stayed in the same general region of the semantic space.
However by 1975, it had begun a transition over to its current meaning \textemdash a shift which accelerated over the years to come.

\begin{figure}[t!]
\begin{subfigure}{0.5\textwidth}
  \centering

  \begin{tabular}{>{\centering}p{0.5\columnwidth}>{\centering}p{0.5\columnwidth}}
  \textbf{\texttt{Sandy}} & \textbf{\texttt{Hurricane}}
  \end{tabular}
  \includegraphics[trim=0 0 0 0, clip=true,width=\textwidth]{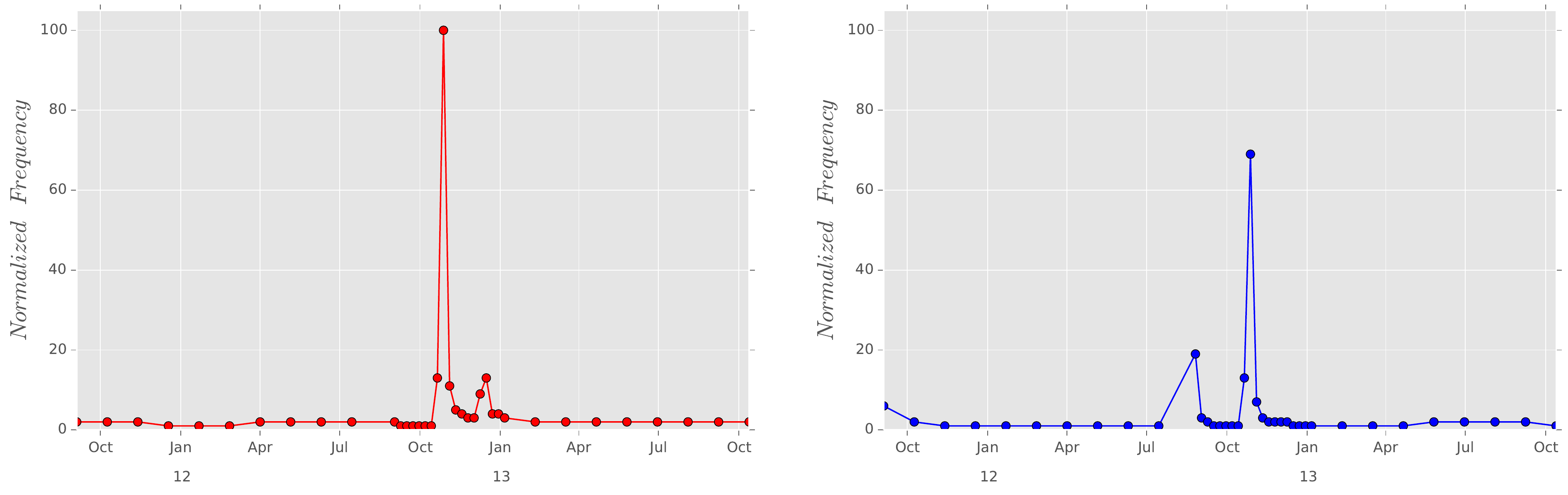}
  \caption{\textbf{Frequency method (Google Trends)}}
  \label{fig:crownjewel_google}
\end{subfigure}%
\\
\\
\begin{subfigure}{0.5\textwidth}
  \centering
  \includegraphics[trim=0 0 0 0, clip=true, width=\textwidth]{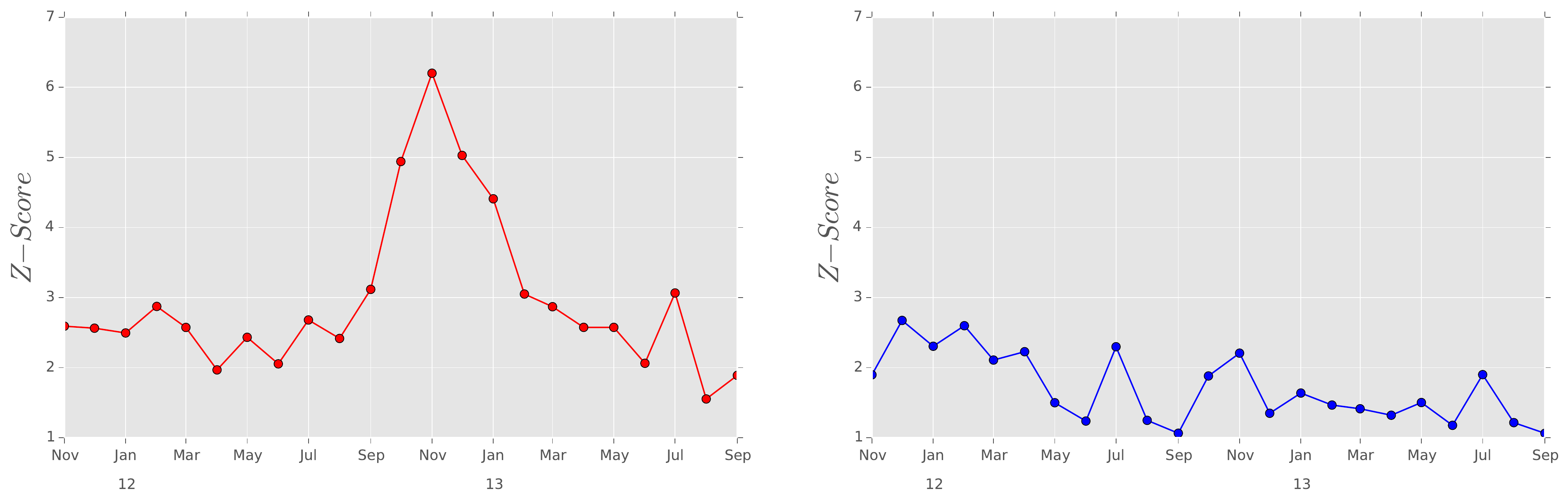}
  \caption{\textbf{Distributional method}}
  \label{fig:crownjewel_our}
\end{subfigure}%
\caption{Comparison between Google Trends and our method. Observe how Google Trends shows spikes
in frequency for both \texttt{Hurricane} (blue) and \texttt{Sandy} (red).
Our method, in contrast, models change in usage and detects that only \texttt{Sandy} changed its meaning and not \texttt{Hurricane}.}
\label{fig:crownjewel}
\end{figure}

The choice of the time series construction method determines the type of information we capture regarding word usage.
The difference between frequency-based approaches and distributional methods is illustrated in Figure \ref{fig:crownjewel}.
Figure \ref{fig:crownjewel_google} shows the frequencies of two words, \texttt{Sandy} (red), and \texttt{Hurricane} (blue) as a percentage of search queries according to Google Trends\footnote{\texttt{http://www.google.com/trends/}}.
Observe the sharp spikes in both words' usage in October 2012, which corresponds to a storm called \texttt{Hurricane Sandy} striking the Atlantic Coast of the United States.
However, only one of those words (\texttt{Sandy}) actually acquired a new meaning.
Indeed, using our distributional method (Figure \ref{fig:crownjewel_our}), we observe that only the word \texttt{Sandy} shifted in meaning where as \texttt{Hurricane} did not.

Our computational approach is scalable, and we demonstrate this by running our method on three large datasets.
Specifically, we investigate linguistic change detection across years of micro-blogging using Twitter, a decade of product reviews using a corpus of movie reviews from Amazon, and a century of written books using the \ngrams.

Despite the fast pace of change of the web content,
our method is able to detect the introduction of new products, movies and books.
This could help semantically aware web applications to better understand user intentions and requests.
Detecting the semantic shift of a word would trigger such applications to apply focused sense disambiguation analysis.


In summary, our contributions are as follows:
\begin{itemize}
\item \textbf{Word Evolution Modeling}: We study three different methods for the statistical modeling of word evolution over time.
We use measures of frequency, part-of-speech tag distribution, and word co-occurrence to construct time series for each word under investigation.(Section \ref{sec:construction})
\item \textbf{Statistical Soundness}: We propose (to our knowledge) the first statistically sound method for linguistic shift detection.  Our approach uses change point detection in time series to assign significance of change scores to each word. (Section \ref{sec:detection})
\item \textbf{Cross-Domain Analysis}: We apply our method on three different domains; books, tweets and online reviews.
Our corpora consists of billions of words and spans several time scales.
We show several interesting instances of semantic change identified by our method. (Section \ref{sec:exps})
\end{itemize}

The rest of the paper is structured as follows.
In Section \ref{sec:def} we define the problem of language shift detection over time.
Then, we outline our proposals to construct time series modeling word evolution in Section \ref{sec:construction}.
Next, in Section \ref{sec:detection}, we describe the method we developed for detecting significant changes in natural language.
We describe the datasets we used in Section \ref{sec:datasets}, and then evaluate our system both qualitatively and quantitatively in Section \ref{sec:exps}.
We follow this with a treatment of related work in Section \ref{sec:related}, and finally conclude with a discussion of the limitations and possible future work in Section \ref{sec:Conclusions}.

\section{Problem Definition}
\label{sec:def}

Our problem is to quantify the linguistic shift in word meaning and usage across time.
Given a temporal corpora $\mathcal{C}$ that is created over a time span $\mathcal{S}$, we divide the corpora into $n$ snapshots $\mathcal{C}_t$ each of period length $P$.
We build a common vocabulary $\mathcal{V}$ by intersecting the word dictionaries that appear in all the snapshots (i.e, we track the same word set across time).
This eliminates trivial examples of word usage shift from words which appear or vanish throughout the corpus.

To model word evolution, we construct a time series $\mathcal{T}(w)$ for each word $w \in \mathcal{V}$.
Each point $\mathcal{T}_t(w)$ corresponds to statistical information extracted from corpus snapshot $\mathcal{C}_t$ that reflects the usage of $w$.
In Section \ref{sec:construction}, we propose several methods to calculate $\mathcal{T}_t(w)$, each varying in the statistical information used to capture $w$'s usage.

Once these time series are constructed, we can quantify the significance of the shift that occurred to the word in its meaning and usage.
Sudden increases or decreases in the time series are indicative of shifts in the word usage. 
Specifically we pose the following questions:
\begin{enumerate}
\item How statically significant is the shift in usage of a word $w$ across time (in $\mathcal{T}(w)$)?.
\item Given that a word has shifted, at what point in time did the change happen?
\end{enumerate}


\section{Time Series Construction}
\label{sec:construction}
Constructing the time series is the first step in quantifying the significance of word change.
Different approaches capture different aspects of word's semantic, syntactic and usage patterns.
In this section, we describe three approaches (Frequency, Syntactic, and Distributional) to building a time series that capture different aspects of word evolution across time.
The choice of time series significantly influences the types of changes we can detect \textemdash a phenomenon which we discuss further in Section \ref{sec:exps}.

\subsection{Frequency Method}
\label{sec:freq}
The most immediate way to detect sequences of discrete events is through their change in frequency.
Frequency based methods are therefore quite popular, and include tools like Google Trends and \ngrams, both of which are used in research to predict economical and public health changes \cite{gtrends,gtrends2}.
Such analysis depends on keyword search over indexed corpora.

\begin{figure}[tb]
\centering
\includegraphics[width=\columnwidth]{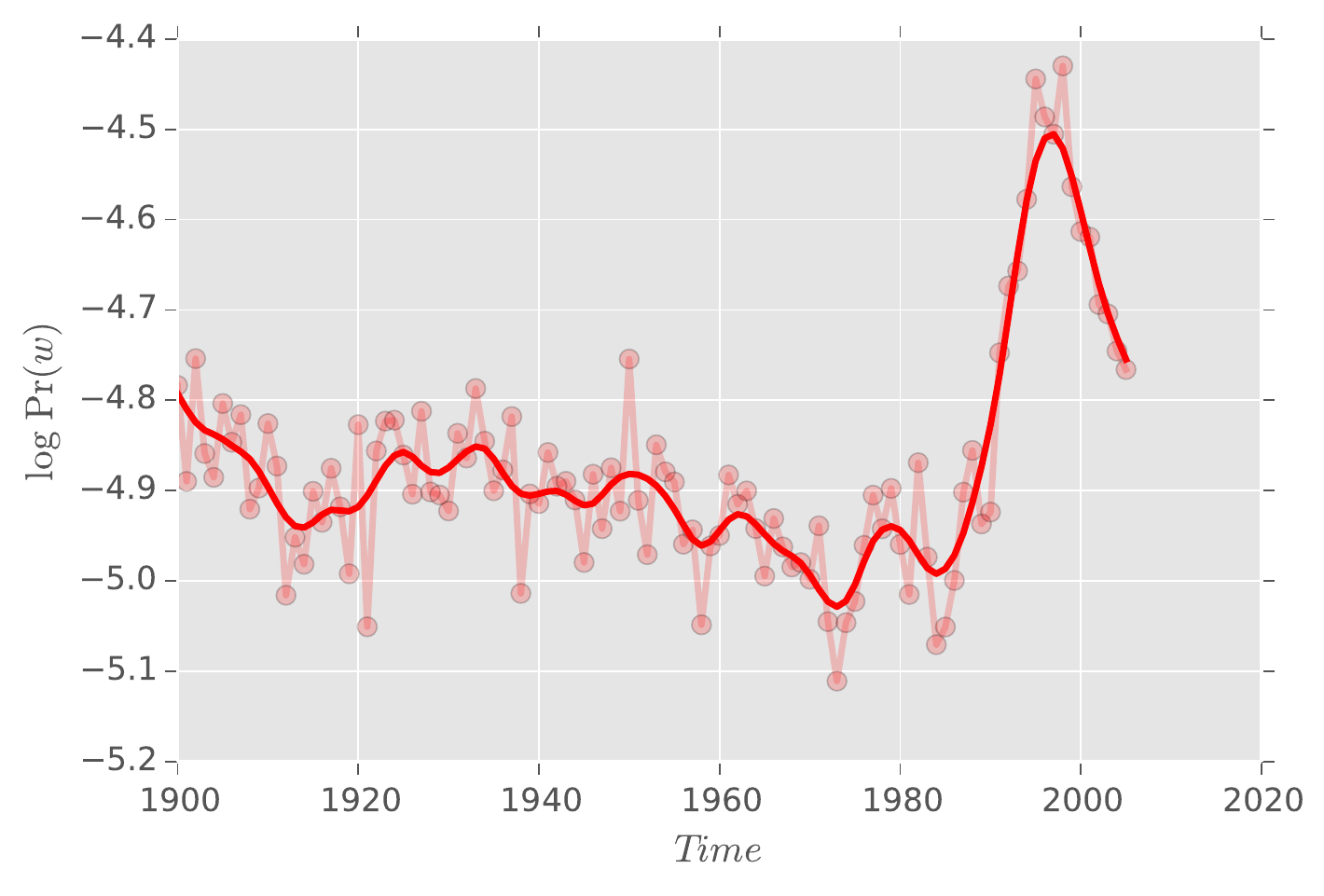}
\caption{Frequency usage of the word \texttt{gay} over time, observe the sudden change in frequency in the late 1980s.}
\label{fig:freq}
\end{figure}

Frequency based methods can capture linguistic shift, as changes in frequency can correspond to words acquiring or losing senses.
Although crude, this method is simple to implement.
We track the change in probability of a word appearing over time.
We calculate for each time snapshot corpus $\mathcal{C}_t$, a unigram language model.
Specifically, we construct the time series for a word $w$ as follows:
\begin{equation}
\mathcal{T}_t(w) = \log \frac{\#(w \in \mathcal{C}_t)}{|\mathcal{C}_t|},
\end{equation}
where $\#(w \in \mathcal{C}_t)$ is the number of occurrences of the word $w$ in corpus snapshot $\mathcal{C}_t$.
An example of the information we capture by tracking word frequencies over time is shown in Figure \ref{fig:freq}.
Observe the sudden jump in late 1980s of the word \texttt{gay} in frequency.

\subsection{Syntactic Method}
While word frequency based metrics are easy to calculate, they are prone to sampling error introduced by bias in domain and genre distribution in the corpus.
Temporal events and popularity of specific entities could spike the word usage frequency without significant shift in its meaning, recall \texttt{Hurricane} in Figure \ref{fig:crownjewel_google}.

\begin{figure}[t!]
  \centering
  \includegraphics[width=1.05\columnwidth]{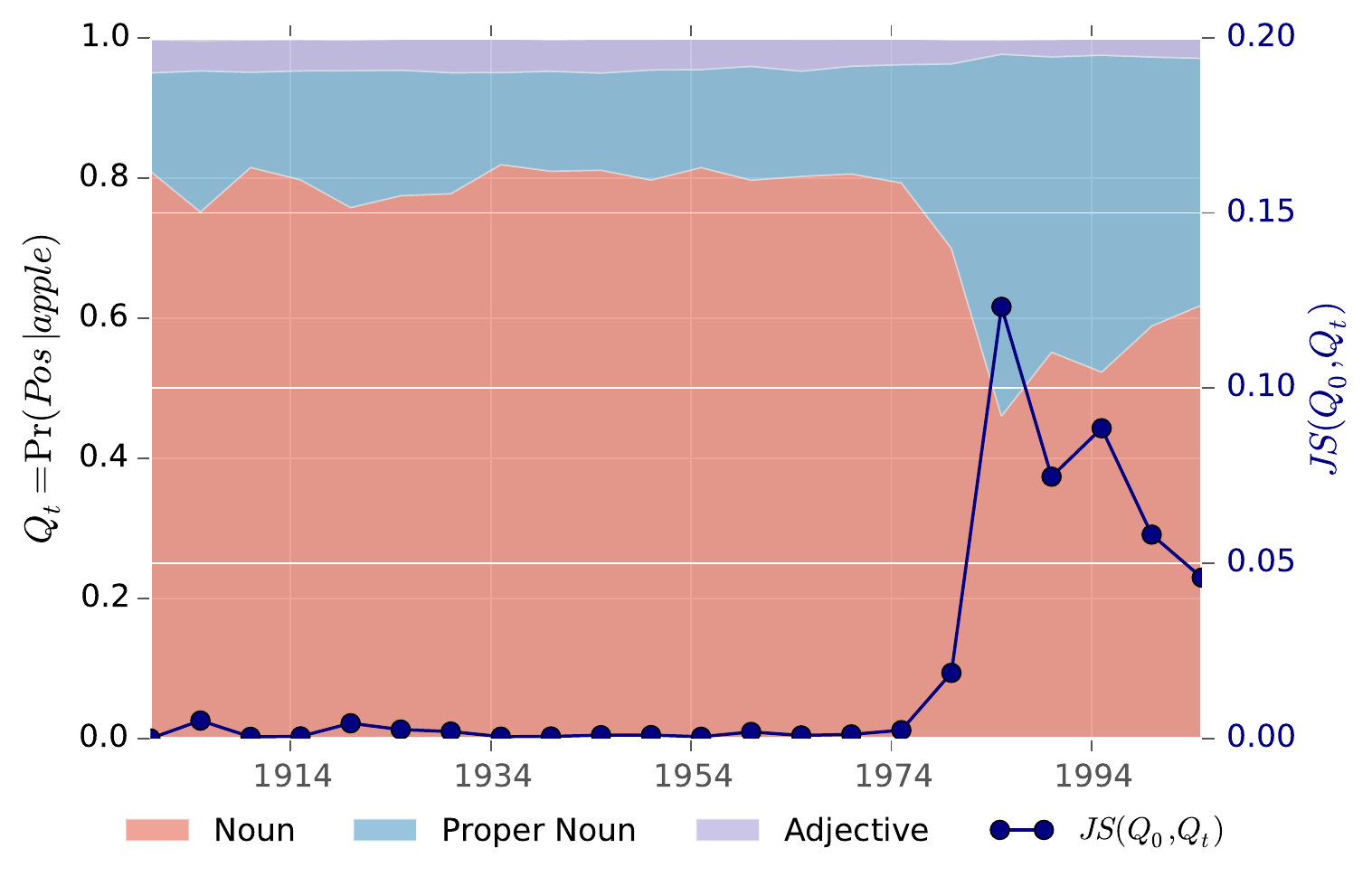}
\caption{Part of speech tag probability distribution of the word \texttt{apple} (stacked area chart).
Observe that the ``Proper Noun" tag has dramatically increased in 1980s.
The same trend is clear from the time series constructed using Jenssen-Shannon Divergence (dark blue line).
}
\label{fig:pos}
\end{figure}

Another approach to detect and quantify significant change in the word usage involves tracking the syntactic functionality it serves.
A word could evolve a new syntactic functionality by acquiring a new part of speech category.
For example, \texttt{apple} used to be only a ``Noun" describing a fruit, but over time it acquired the new part of speech ``Proper Noun" to indicate the new sense describing a technical company (Figure \ref{fig:pos}).
To leverage this syntactic knowledge, we annotate our corpus with part of speech (POS) tags.
Then we calculate the probability distribution of part of speech tags $Q_t$ given the word $w$ and time snapshot $t$ as follows: $Q_t = \Pr_{X \overset{}{\sim} \text{POS Tags}}(X | w, \mathcal{C}_t)$.
We consider the POS tag distribution at $t=0$ to be the initial distribution $Q_0$.
To quantify the temporal change between two time snapshots corpora, for a specific word $w$, we calculate the divergence between the POS distributions in both snapshots.

Specifically, we construct the time series as follows:

\begin{align}
\mathcal{T}_t(w) =& \;\;\; JSD(Q_0, Q_t)
\end{align}
where JSD is the Jenssen-Shannon divergence \cite{lin1991divergence}.

Figure \ref{fig:pos} shows that the JS divergence (dark blue line) reflects the change in the distribution of the part of speech tags given the word \texttt{apple}.
In 1980s,  the ``Proper Noun" tag (blue area) increased dramatically due to the rise of \texttt{Apple Computer Inc.}, the popular consumer electronics company.

\subsection{Distributional Method}
Semantic shifts are not restricted to changes to part of speech.
For example, consider the word \texttt{mouse}. In the 1970s it acquired a new sense of ``computer input device", but did not change its part of speech categorization (since both senses are nouns).
To detect such subtle semantic changes, we need to infer deeper cues from the contexts a word is used in.

The distributional hypothesis states that words appearing in similar contexts are semantically similar \cite{firth1961papers}.
Distributional methods learn a semantic space that maps words to continuous vector space $\mathbb{R}^d$, where $d$ is the dimension of the vector space.
Thus, vector representations of words appearing in similar contexts will be close to each other.
Recent developments in representation learning (\emph{deep learning}) \cite{learningrepresentations} have enabled the scalable learning of such models.
We use a variation of these models \cite{skipgram} to learn word vector representation (\emph{word embeddings}) that we track across time.

Specifically, we seek to learn a temporal word embedding $\phi_t:  \mathcal{V}, \mathcal{C}_t \mapsto \mathbb{R}^d$.
Once we learn a representation of a specific word for each time snapshot corpus, we track the changes of the representation across the embedding space to quantify the meaning shift of the word (as shown in Figure \ref{fig:vis_embeddings}).

In this section we present our distributional approach in detail. Specially we discuss the learning of word embeddings, the aligning of embedding spaces across different time snapshots to a joint embedding space, and the utilization of a word's displacement through this semantic space to construct a distributional time series.

\begin{figure}[t]
\centering
\includegraphics[width=\columnwidth]{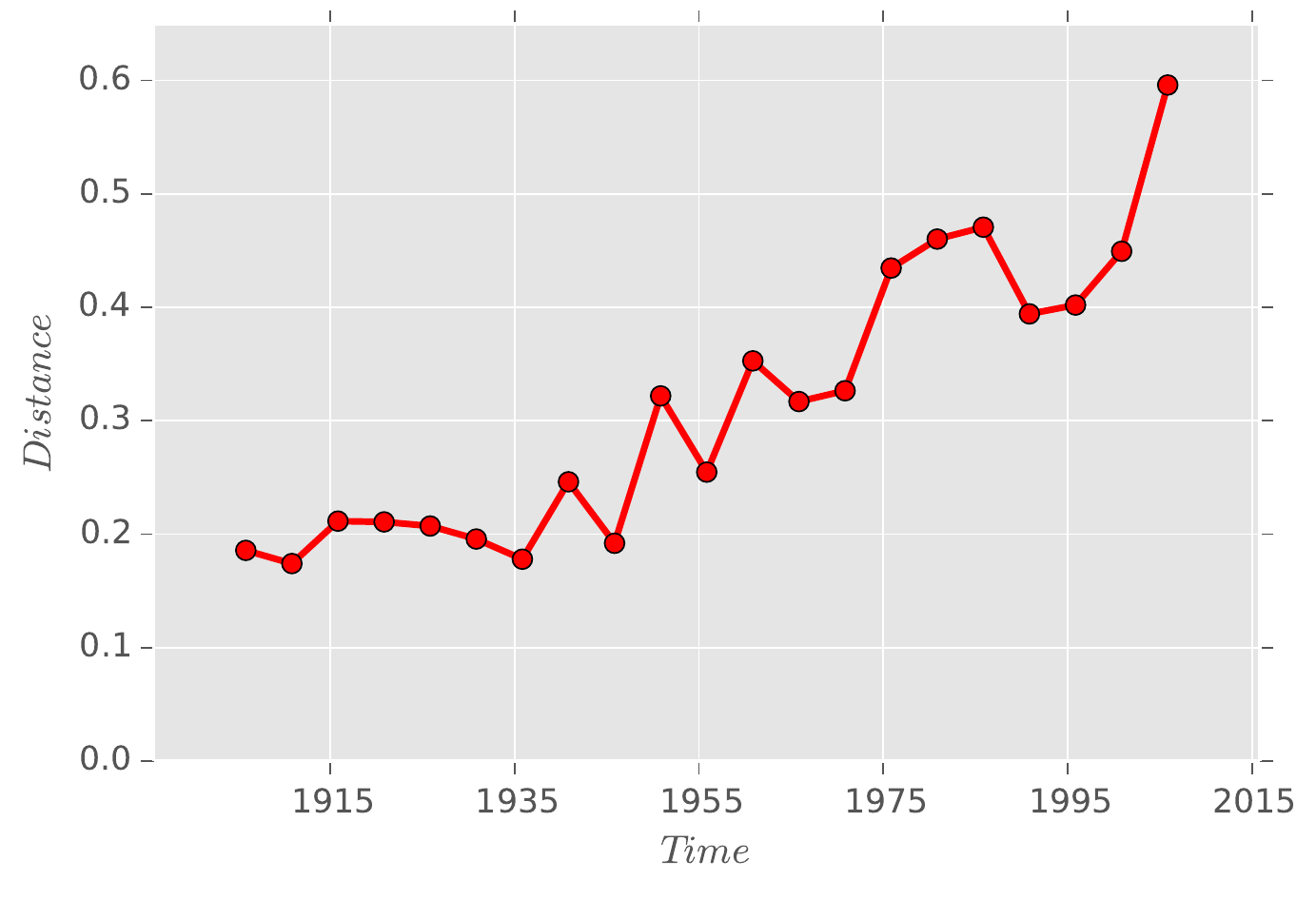}
\caption{Distributional time series for the word \texttt{tape} over time using word embeddings. Observe the change of behavior starting in the 1950s, which is quite apparent by the 1970s.}
\label{fig:embeddings}
\end{figure}

\subsubsection{Learning embeddings}
Given a time snapshot $\mathcal{C}_t$ of the corpus, our goal is to learn $\phi_t$ over $\mathcal{V}$ using neural language models.
At the beginning of the training process, the words vector representations are randomly initialized.
The training objective is to maximize the probability of the words appearing in the context of word $w_i$.
Specifically, given the vector representation $\mathbf{w}_i$ of a word $w_i$ ($\mathbf{w}_i = \phi_t(w_i)$),
we seek to maximize the probability of $w_j$ through the following equation:
\begin{align}
\Pr(w_j \mid \mathbf{w}_i) &= \frac{\exp{(\mathbf{w}_j^T\mathbf{w}_i)}}
{\sum\limits_{w_k \in \mathcal{V}} \exp{(\mathbf{w}_k^T\mathbf{w}_i)}}
\label{eq:prob1}
\end{align}
In a single epoch, we iterate over each word occurrence in the time snapshot $\mathcal{C}_t$ to minimize the negative log-likelihood of the context words.
Context words are the words appearing to the left or right of $w_i$ within a window of size $m$ (Equation \ref{eq:prob2}).
\begin{align}
J &= \sum_{w_i \in \mathcal{C}_t} \sum_{\substack{j=i-m\\j!=i}}^{i+m} - \log \Pr(w_j \mid \mathbf{w}_i)
\label{eq:prob2}
\end{align}

Notice that the normalization factor that appears in Equation \ref{eq:prob1} is not feasible to calculate if $|\mathcal{V}|$ is too large.
To approximate this probability, we map the problem from a classification of 1-out-of-$\mathcal{V}$ words to a hierarchical classification problem \cite{hsm1,hsm2}.
This reduces the cost of calculating the normalization factor from $\mathcal{O}(|\mathcal{V}|)$ to $\mathcal{O}(\log |\mathcal{V}|)$.

We optimize the model parameters using stochastic gradient descent \cite{sgd}, as follows:
\begin{equation}
\phi_t(w_i) = \phi_t(w_i) - \alpha \times \frac{\partial{J}}{\partial\phi_t(w_i)},
\end{equation}
where $\alpha$ is the learning rate.
We calculate the derivatives of the model using the back-propagation algorithm. \cite{backpropagation}.
We use the following measure of training convergence:

\begin{equation}
\rho = \frac{1}{|\mathcal{V}|}\sum_{w \in \mathcal{V}} \frac{{\phi^k}^T(w) \phi^{k+1}(w)} {\norm{\phi^k(w)}_2  \norm{\phi^{k+1}(w)}_2},
\end{equation}
where $\phi^k$ is the model parameters after epoch $k$.
We calculate $\rho$ after each epoch and stop the training if $\rho \le 1.0^{-4}$.
After training stops, we normalize word embeddings by their $L_2$ norm, which forces all words to be represented by unit vectors.

In our experiments, we use gensim implementation of skipgram models\footnote{\url{https://github.com/piskvorky/gensim}}.
We set the context window size $m$ to 10 unless otherwise stated.
We choose the size of the word embedding space dimension $d$ to be 200.
To speed up the training, we subsample the frequent words by the ratio $10^{-5}$ \cite{phrases}.

\subsubsection{Aligning Embeddings}
\label{sec:align}
Having trained temporal word embeddings for each time snapshot $\mathcal{C}_t$, we must now align the embeddings so that all the embeddings are in one unified co-ordinate system. This enables us to characterize the change between them.
This process is complicated by the stochastic nature of our training, which implies that models trained on exactly the same data could produce vector spaces where words have the same nearest neighbors but not with the same coordinates.
The alignment problem is exacerbated by actual changes in the distributional nature of words in each snapshot.

To aid the alignment process, we make two simplifying assumptions:
First, we assume that the spaces are equivalent under a linear transformation.
Second, we assume that the meaning of most words did not shift over time, and therefore, their local structure is preserved.
Based on these assumptions, observe that when the alignment model fails to align a word properly, it is possibly indicative of a linguistic shift.



Specifically, we define the set of $k$ nearest words in the embedding space $\phi_t$ to a word $w$ to be $k$-NN$(\phi_t(w))$.
We seek to learn a linear transformation $\mathbf{W}_{t'\mapsto t}(w) \in \mathbb{R}^{d\times d}$ that maps a word from $\phi_t$ to $\phi_{t'}$ by solving the following optimization problem:

\begin{equation}
\underset{t'\mapsto t}{\mathbf{W}}(w) = \argmin\limits_{\mathbf{W}} \sum_{\underset{k\text{-NN}(\phi_{t'}(w))}{w_i \in}} \norm{\phi_{t'}(w_i)\mathbf{W} - \phi_t(w_i)}_2^2,
\label{eq:warp}
\end{equation}
which is equivalent to a piecewise linear regression model.

\begin{figure*}[t!]
  \centering
  \includegraphics[width=2\columnwidth]{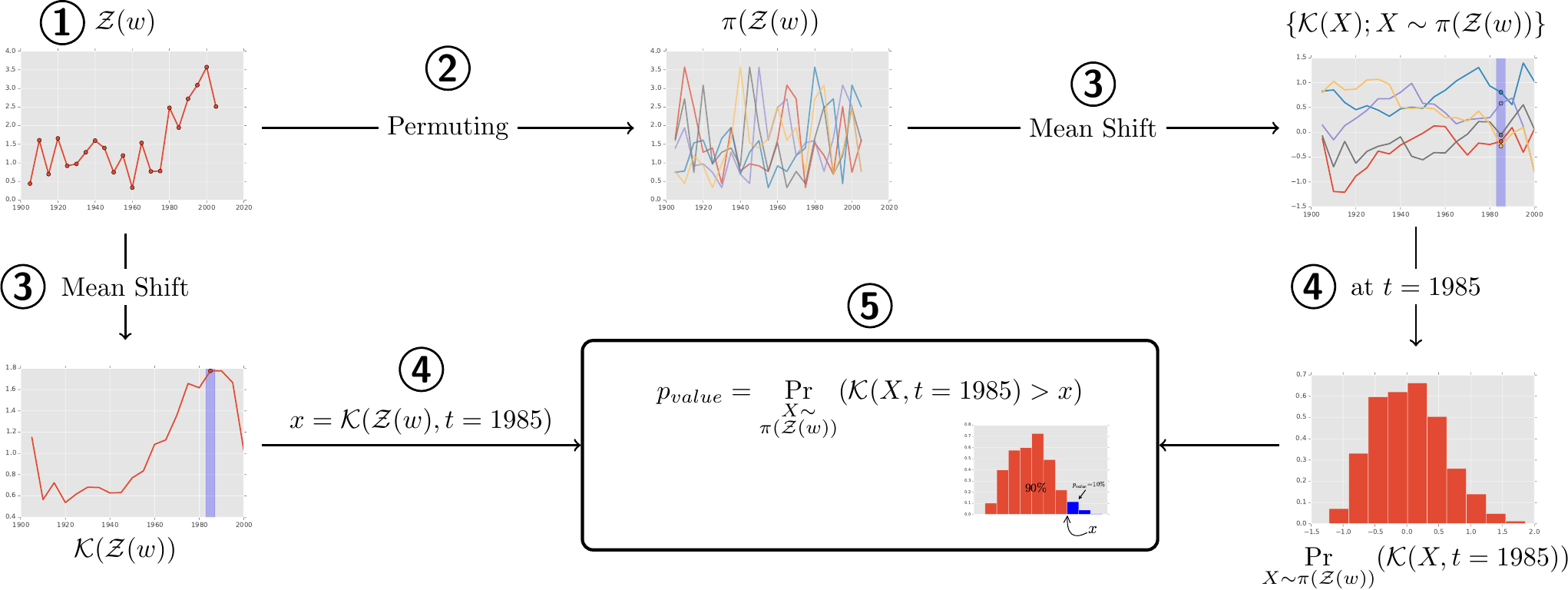}
  \vspace{0.5em}
  \caption{Our change point detection algorithm.
   In Step \ding{172}, we normalize the given time series $\mathcal{T}(w)$ to produce $\mathcal{Z}(w)$.
   Next, we shuffle the time series points producing the set  $\pi(\mathcal{Z}(w))$ (Step \ding{173}).
   Then, we apply the mean shift transformation ($\mathcal{K}$) on both the original normalized time series $\mathcal{Z}(w)$ and the permuted set (Step \ding{174}).
   In Step \ding{175}, we calculate the probability distribution of the mean shifts possible given a specific time ($t=1985$) over the bootstrapped samples.
   Finally, we compare the observed value in $\mathcal{K}(\mathcal{Z}(w))$ to the probability distribution of possible values to calculate the \pvalue\ which determines the statistical significance of the observed time series shift (Step \ding{176}).
 }
  \label{fig:algo_vis}
\end{figure*}%

\subsubsection{Time series construction}
To track the shift of word position across time, we align all embeddings spaces to the embedding space of the initial time snapshot $\phi_0$ using a linear mapping (Eq.\ \ref{eq:warp}).
This unification of coordinate system allows us to compare relative displacements that occurred to words across different time periods.

To capture linguistic shift, we construct our distributional time series by calculating the distance in the embedding space between $\phi_t(w)\mathbf{W}_{t\mapsto 0}(w)$ and $\phi_0(w)$ as the following:

\begin{equation}
\mathcal{T}_t(w)= 1 - \frac{(\phi_t(w) \mathbf{W}_{t\mapsto 0}(w))^T \phi_0(w)}{\norm{\phi_t(w)\mathbf{W}_{t\mapsto 0}(w)}_2\norm{\phi_0(w)}}_2
\end{equation}

Figure \ref{fig:embeddings} shows the time series obtained using word embeddings for \texttt{tape}, which underwent a semantic change in the 1950's with the introduction of magnetic tape recorders.
As such recorders grew in popularity, the change becomes more pronounced, until it is quite apparent by the 1970s.


\section{Change Point Detection}
\label{sec:detection}
Given a time series of a word $\mathcal{T}(w)$, constructed using one of the methods discussed in Section \ref{sec:construction}, we seek to determine whether the word changed significantly, and if so estimate the change point.

\begin{algorithm}[t!]
    \caption{\small \textsc{Change Point Detection} ($\mathcal{T}(w)$, $B$, $\gamma$)}
    \label{alg:deepshift}
    \begin{algorithmic}[1]
        \REQUIRE $\mathcal{T}(w)$: Time series for the word $w$, $B$: Number of bootstrap samples, $\gamma$: Z-Score threshold
        \ENSURE $ECP$: Estimated change point, \pvalue: Significance score.
        \STATEx // Preprocessing
        \STATE $Z(w) \gets $ Normalize $\mathcal{T}(w)$.
        \STATE Compute mean shift series $\mathcal{K}(Z(w))$ \label{lst:line:ms}
        \STATEx // Bootstrapping
		\STATE $BS \gets \emptyset $ \Comment {Bootstrapped samples} \label{lst:line:bsbegin}
        \REPEAT
        \STATE Draw $P$ from $\pi(\mathcal{Z}(w))$
        \STATE $BS \gets BS \cup P$
        \UNTIL{|$BS$| = $B$}
        \FOR{$i\gets 1, n$}
            \STATE $p$-value$(w, i) \gets \frac{1}{B} \sum_{P \in BS} [\mathcal{K}_i(P) > \mathcal{K}_i(Z(w))]$
        \ENDFOR \label{lst:line:bsend}
        \STATEx // Change Point Detection
        \STATE $C \gets {\{j | j \in [1,n]\ and\ Z_{j}(w)>= \gamma \}}$ \label{lst:line:ecpbegin}
        \STATE \pvalue\ $\gets \min_{j \in C} p$-value$(w,j)$
        \STATE $ECP \gets \argmin_{j \in C} p$-value$(w,j)$
        \STATE \textbf{return} $p$-value, $ECP$  \label{lst:line:ecpend}
    \end{algorithmic}
\end{algorithm}

There exists an extensive body of work on change point detection in time series \cite{Taylor00change-pointanalysis:, Basseville:1993:DAC:151741, adams-mackay-2007a}.
Our approach models the time series based on the \emph{Mean Shift} model described in \cite{Taylor00change-pointanalysis:}.
First, our method recognizes that language exhibits a general stochastic drift.
We account for this by first normalizing the time series for each word.
Our method then attempts to detect a shift in the mean of the time series using a variant of mean shift algorithms for change point analysis.
We outline our method in Algorithm \ref{alg:deepshift} and describe it below.
We also illustrate key aspects of the method in Figure \ref{fig:algo_vis}.

Given a time series of a word $\mathcal{T}(w)$, we first normalize the time series.
We calculate the mean $\mu_i = \frac{1}{|\mathcal{V}|}\sum_{w \in \mathcal{V}}\mathcal{T}_i(w)$ and standard deviation $Var_i = \frac{1}{|\mathcal{V}|}\sum_{w \in \mathcal{V}}(\mathcal{T}_i(w) - \mu_i)^2$ across all words.
Then, we transform the time series into a $Z-Score$ series as follows:
\begin{equation}
\label{eq:norm}
\mathcal{Z}_i(w) = \frac{\mathcal{T}_i(w) - \mu_i}{Var_i}
\end{equation}
where $\mathcal{Z}_i(w)$ is the $z$-score of the time series for the word $w$ at time snapshot $i$.

We model the time series $\mathcal{Z}(w)$ by a \emph{Mean shift model} \cite{Taylor00change-pointanalysis:}.
Let $S$ = $\mathcal{Z}_1(w),\mathcal{Z}_2(w),\dots,\mathcal{Z}_n(w)$ represent the time series.
We model $\mathcal{S}$ to be an output of a stochastic process where each $\mathcal{S}_i$ can be described as $\mathcal{S}_i = {\mu}_i + {\epsilon}_i$ where ${\mu}_i$ is the mean and ${\epsilon}_i$ is the random error at time $i$.
We also assume that the errors ${\epsilon}_i$ are independent with mean $0$. Generally ${\mu}_i$ = ${\mu}_{i-1}$ except for a few points which are \emph{change points}.

Based on the above model, we define the mean shift of a general time series $S$ as follows:
\begin{equation}
\label{eq:ms}
\mathcal{K}(S) = \frac{1}{l-j} \sum_{k=j+1}^{l} S_k - \frac{1}{j} \sum_{k=1}^{j} S_k
\end{equation}
This corresponds to calculating the shift in mean between two parts of the time series pivoted at time point $j$.
Change points can be thus identified by detecting significant shifts in the mean. \footnote{This is similar to the CUSUM based approach used for detecting change points which is also based on mean shift model.}

Given a normalized time series $\mathcal{Z}(w)$, we then compute the mean shift series $\mathcal{K}(\mathcal{Z}(w))$ (Line \ref{lst:line:ms}).
To estimate the statistical significance of observing a mean shift at time point $j$, we use bootstrapping \cite{efron:bootstrap} (see Figure \ref{fig:algo_vis} and Lines \ref{lst:line:bsbegin}-\ref{lst:line:bsend}) under the null hypothesis that there is no change in the mean.
In particular, we establish statistical significance by first obtaining $B$(typically $B=1000$) bootstrap samples obtained by permuting $\mathcal{Z}(w)$ (Lines \ref{lst:line:bsbegin}-\ref{lst:line:bsend}). Second, for each bootstrap sample P, we calculate $\mathcal{K}(P)$ to yield its corresponding bootstrap statistic and we estimate the statistical significance (\pvalue) of observing the mean shift at time $i$ compared to the $NULL$ distribution(Lines 8-10).
Finally , we estimate the change point by considering the time point $j$ with the minimum \pvalue\ score (described in \cite{taylor:changepoint}). While this method does detect significant changes in the mean of the time series, observe that it does not account for the magnitude of the change in terms of Z-Scores.
We extend this approach to obtain words that changed significantly compared to other words, by considering only those time points where the Z-Score exceeds a user-defined threshold $\gamma$ (we typically set $\gamma$ to 1.75). We then estimate the change point as the time point with the minimum \pvalue\ exactly as outlined before (Lines \ref{lst:line:ecpbegin}-\ref{lst:line:ecpend}).

\section{Datasets}
\label{sec:datasets}

Here we report the details of the three datasets that we consider -  years of micro-blogging from Twitter, a decade of movie reviews from Amazon, and a century of written books using the \ngrams.
Table \ref{tab:datasets} shows a summary of three different datasets spanning different modes of expression on the Internet: books, online forum and a micro-blogs.

\paragraph{The \ngrams} The \ngrams\ project enables the analysis of cultural, social and linguistic trends.
It contains the frequency of short phrases of text (\emph{ngrams}) that were extracted from books written in eight languages over five centuries \cite{Michel:ngramscorpus}.
These ngrams vary in size (1-5) grams. We use the 5-gram phrases which restrict our context window size $m$ to 5.
Here, we show a sample of 5-grams we used:
\vspace{-0.1in}
\begin{framed}
\texttt{\textbullet \; thousand pounds less then nothing \\
\indent
\textbullet \; to communicate to each other
}
\end{framed}
\vspace{-0.1in}
We focus on the time span from $1900-2005$, and set the time snapshot period to 5 years($21$ points).
We obtain the POS Distribution of each word in the above time range by using the Google Syntactic Ngrams dataset \cite{syntacticngrams,syntactic2,syntactic3}.

\paragraph{Amazon Movie Reviews}
Amazon Movie Reviews dataset consists of movie reviews from Amazon.
This data spans August 1997 to October 2012($13$ time points), including all ~8 million reviews.
However, we consider the time period starting from $2000$ as the number of reviews from earlier years is considerably small.
Each review includes product and user information, ratings, and a plain-text review.
A sample review text is shown below:
\vspace{-0.1in}
\begin{framed}
\noindent
\texttt{This movie has it all.Drama, action, amazing battle scenes - the best I've ever seen.It's definitely a must see.}
\end{framed}
\vspace{-0.1in}

\begin{table}[t!]
\begin{tabular}{l|lll}
 & Google Ngrams& Amazon & Twitter \\ \hline
Span (years) & 105  & 12 & 2  \\
Period & 5 years & 1 year & 1 month \\
\# words & $\sim$$10^9$ & $\sim$$9.9 \times 10^8$ & $\sim$$10^9$ \\
$|\mathcal{V}|$ & $\sim$50K & $\sim$50K & $\sim$100K\\
\# documents & $\sim$$7.5 \times 10^8$ & $8.\times 10^6$ & $\sim$$10^8$ \\
Domain & Books & Movie & Micro \\
       &  & Reviews & Blogging \\
\end{tabular}
\caption{Summary of our datasets}
\label{tab:datasets}
\end{table}

\paragraph{Twitter Data} This dataset consists of a sample of that spans 24 months starting from September 2011 to October 2013.
Each Tweet includes the Tweet ID, Tweet and the geo-location if available.
A sample Tweet text is shown below:
\vspace{-0.1in}
\begin{framed}
\noindent
\texttt{I hope sandy doesn't rip the roof off the pool while we're swimming ...}
\end{framed}
\vspace{-0.1in}

\section{Experiments}
\label{sec:exps}
In this section, we apply our method for each dataset presented in Section \ref{sec:datasets} and identify words that have changed usage over time.
We describe the results of our experiments below.

\begin{table*}[t!]
\centering
\begin{tabular}{m{2cm}|m{3cm} m{3cm} m{3cm} m{3cm}}
\textbf{Word} & \multicolumn{3}{c}{\textbf{Time Series}} &\textbf{\pvalue} \\
 & & & & \\
 & \textbf{\freq} & \textbf{\syn} & \textbf{\dist} & \\ \hline
 & & & & \\
\texttt{transmitted} & \includegraphics[scale=0.20]{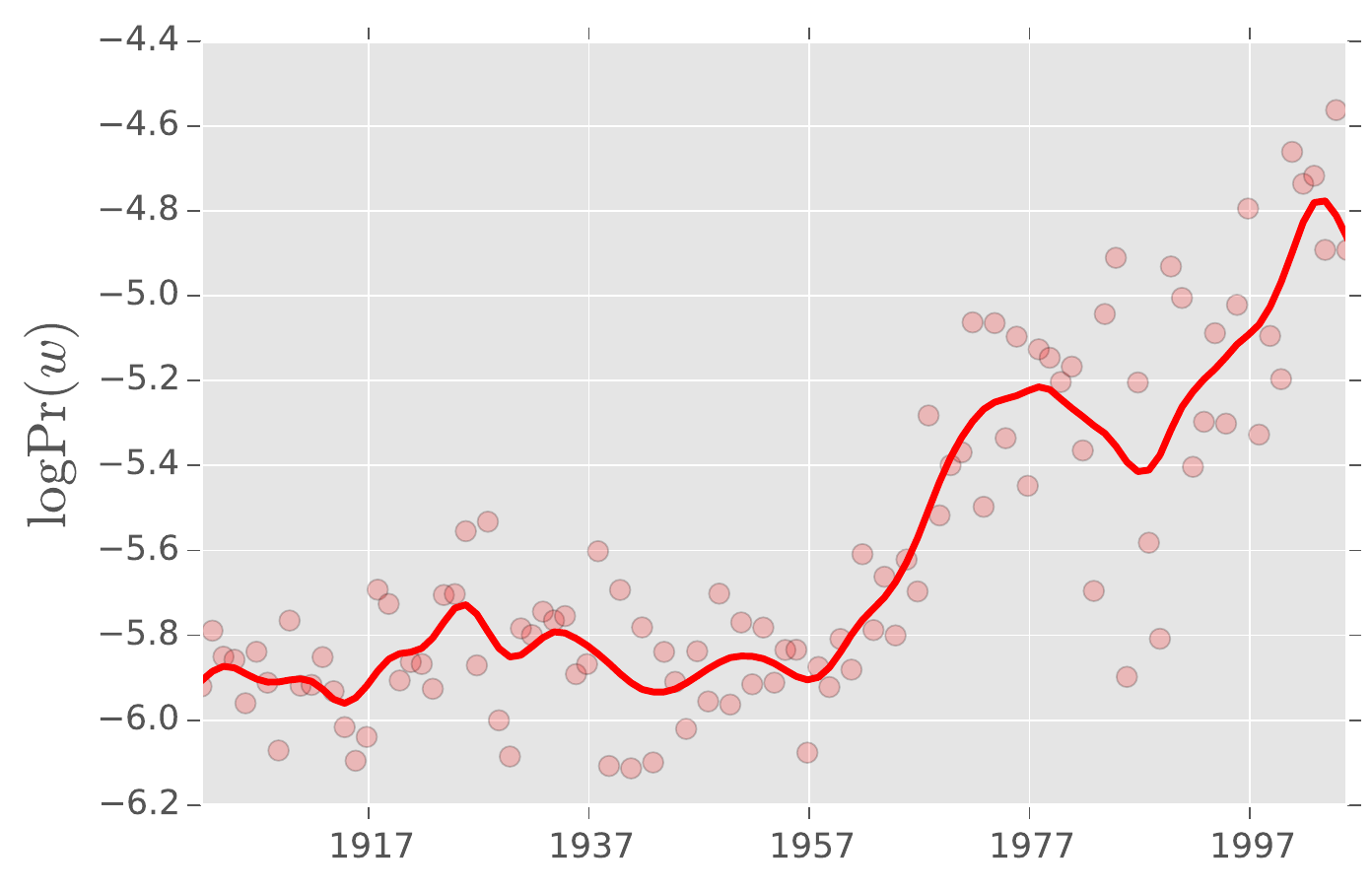} & \includegraphics[scale=0.20]{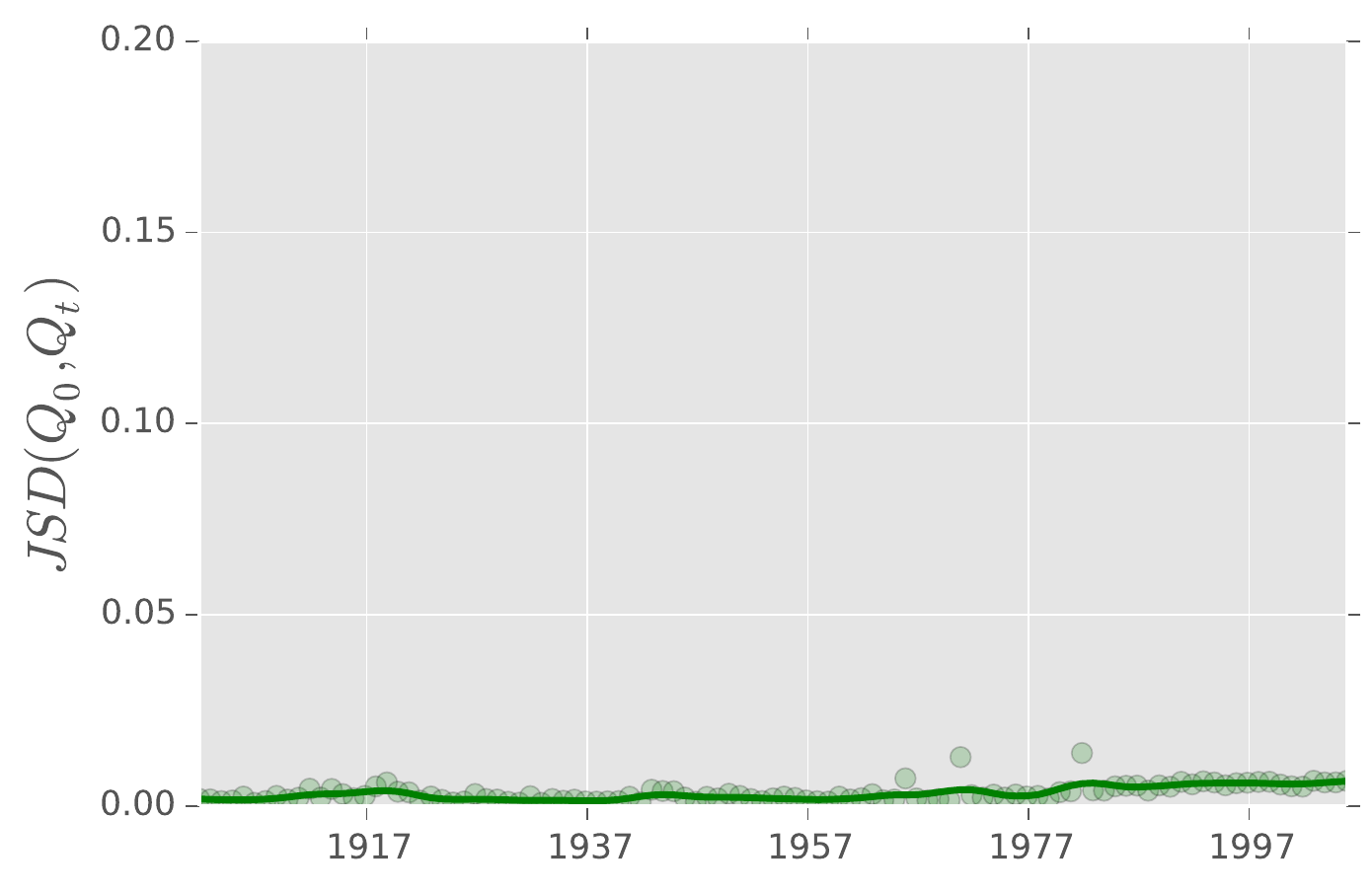}  & \includegraphics[scale=0.20]{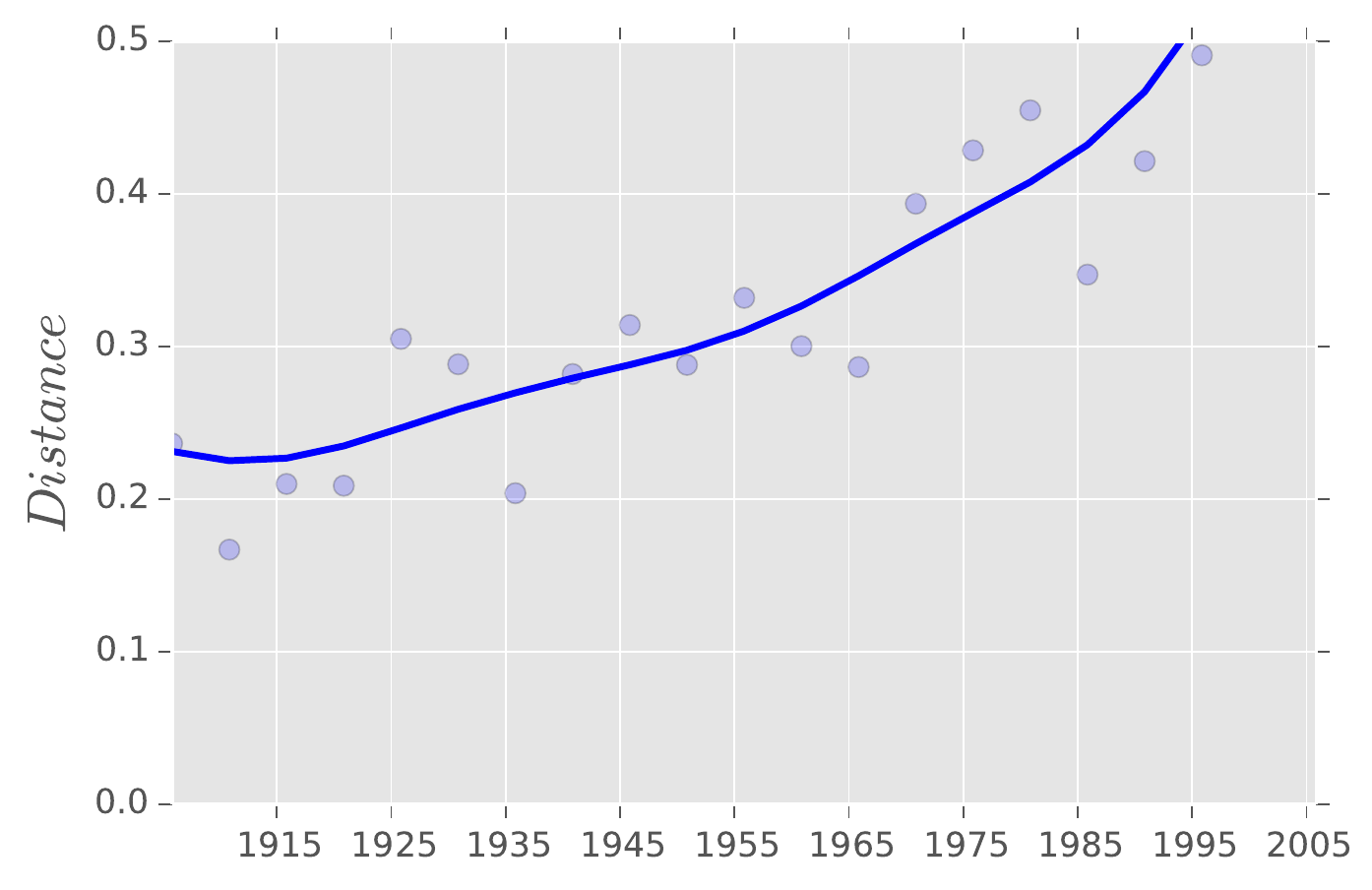}  & \includegraphics[scale=0.20]{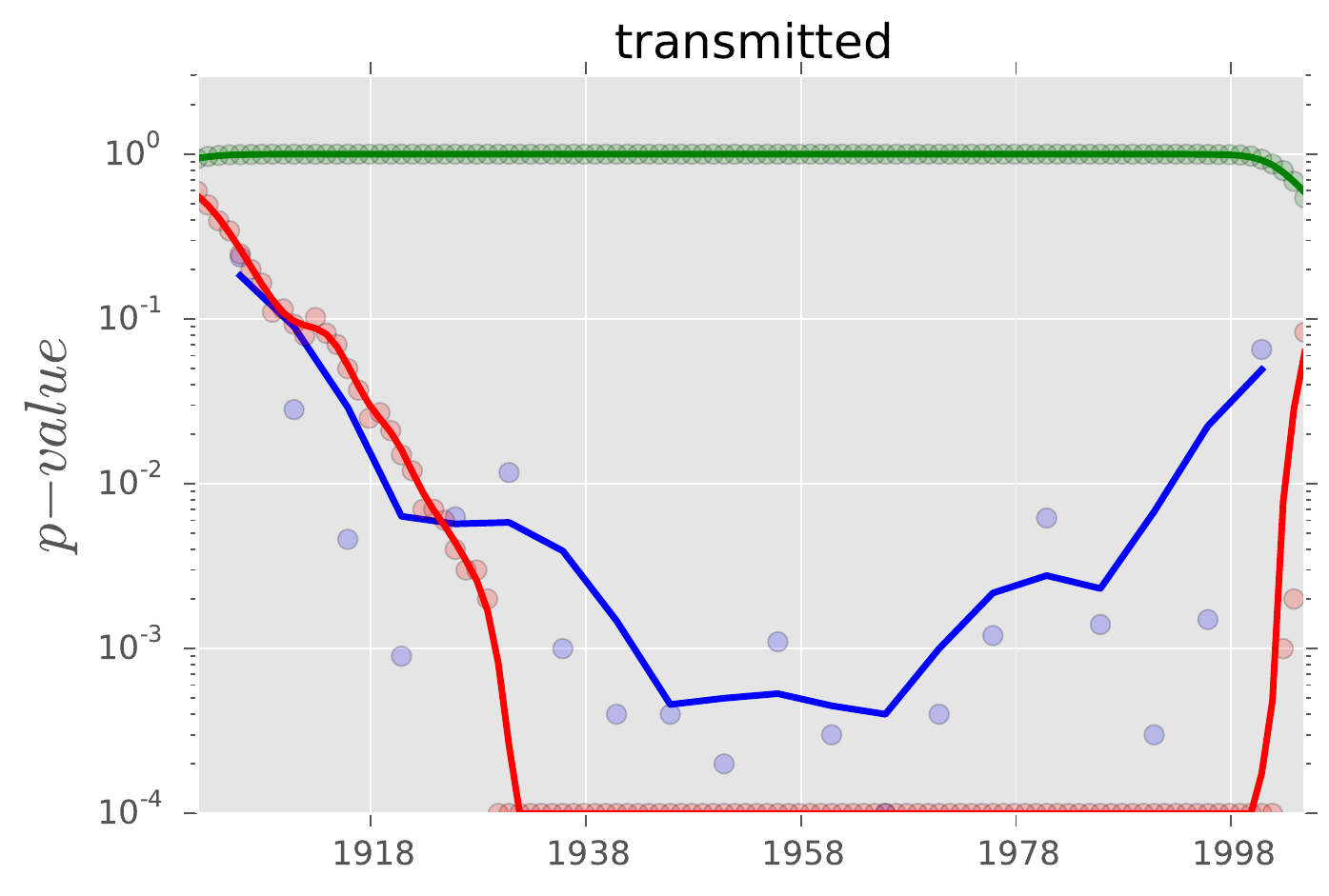} \\
\texttt{bitch} & \includegraphics[scale=0.20]{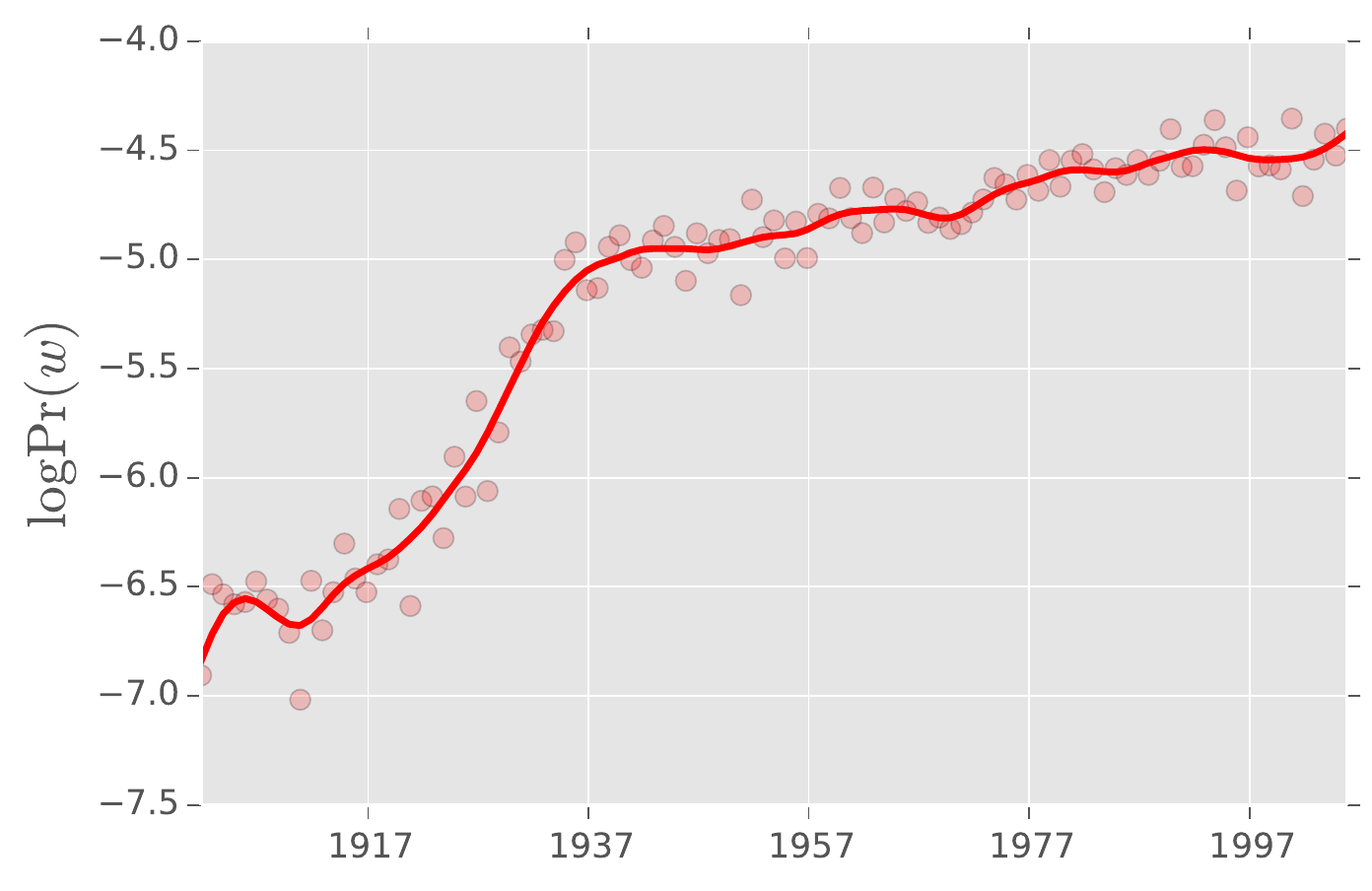} & \includegraphics[scale=0.20]{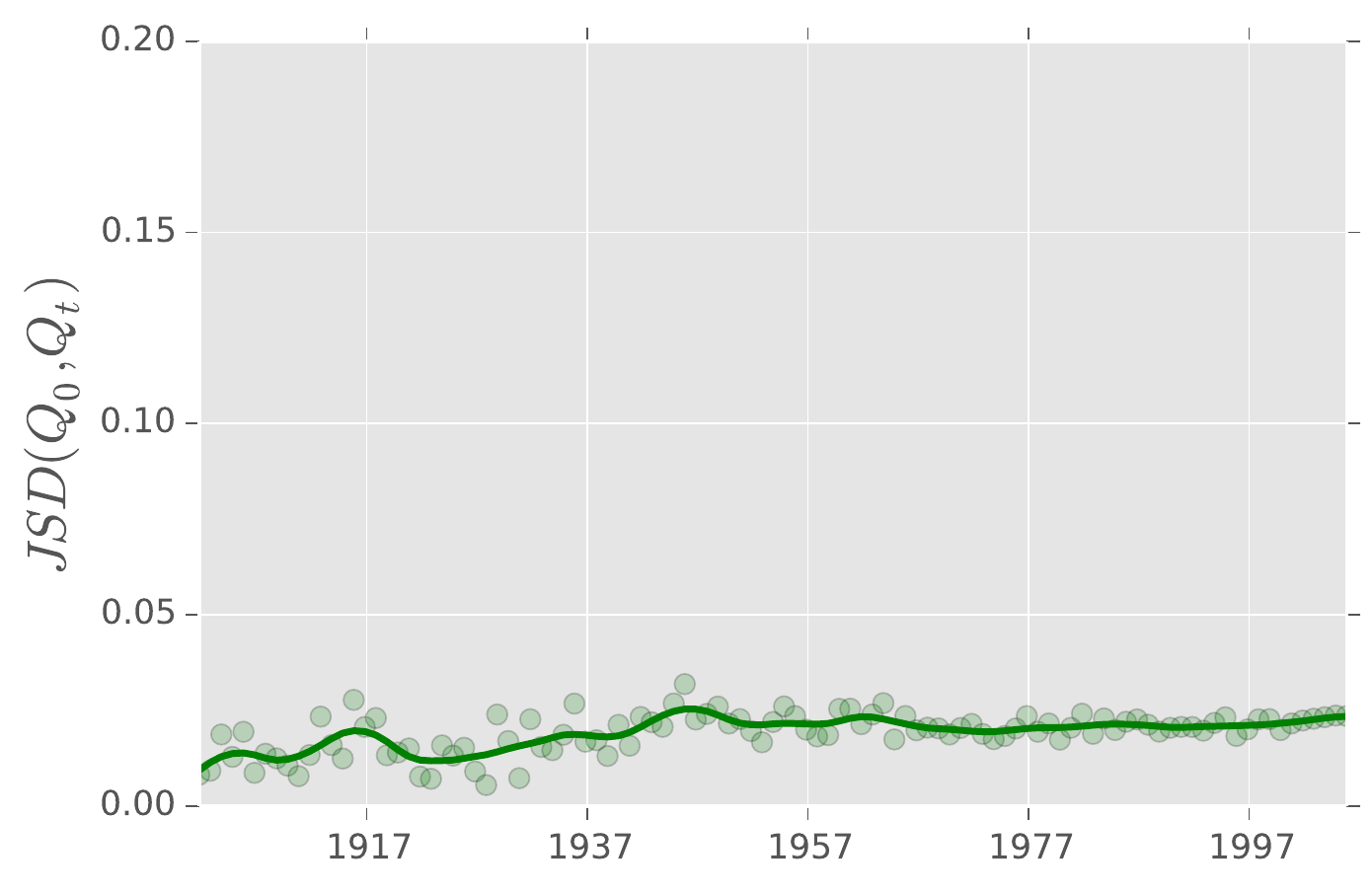}  &
\includegraphics[scale=0.20]{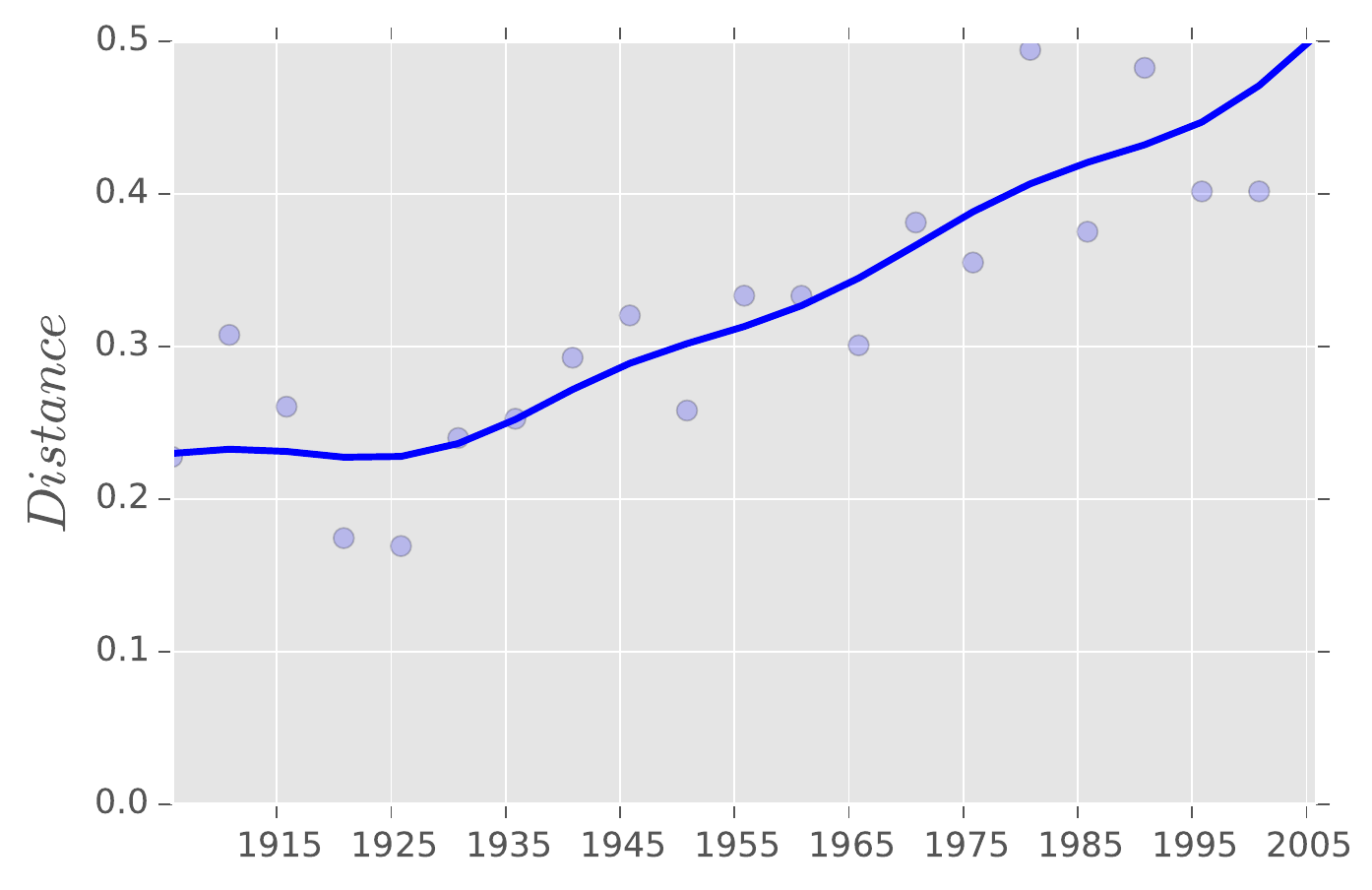} & \includegraphics[scale=0.20]{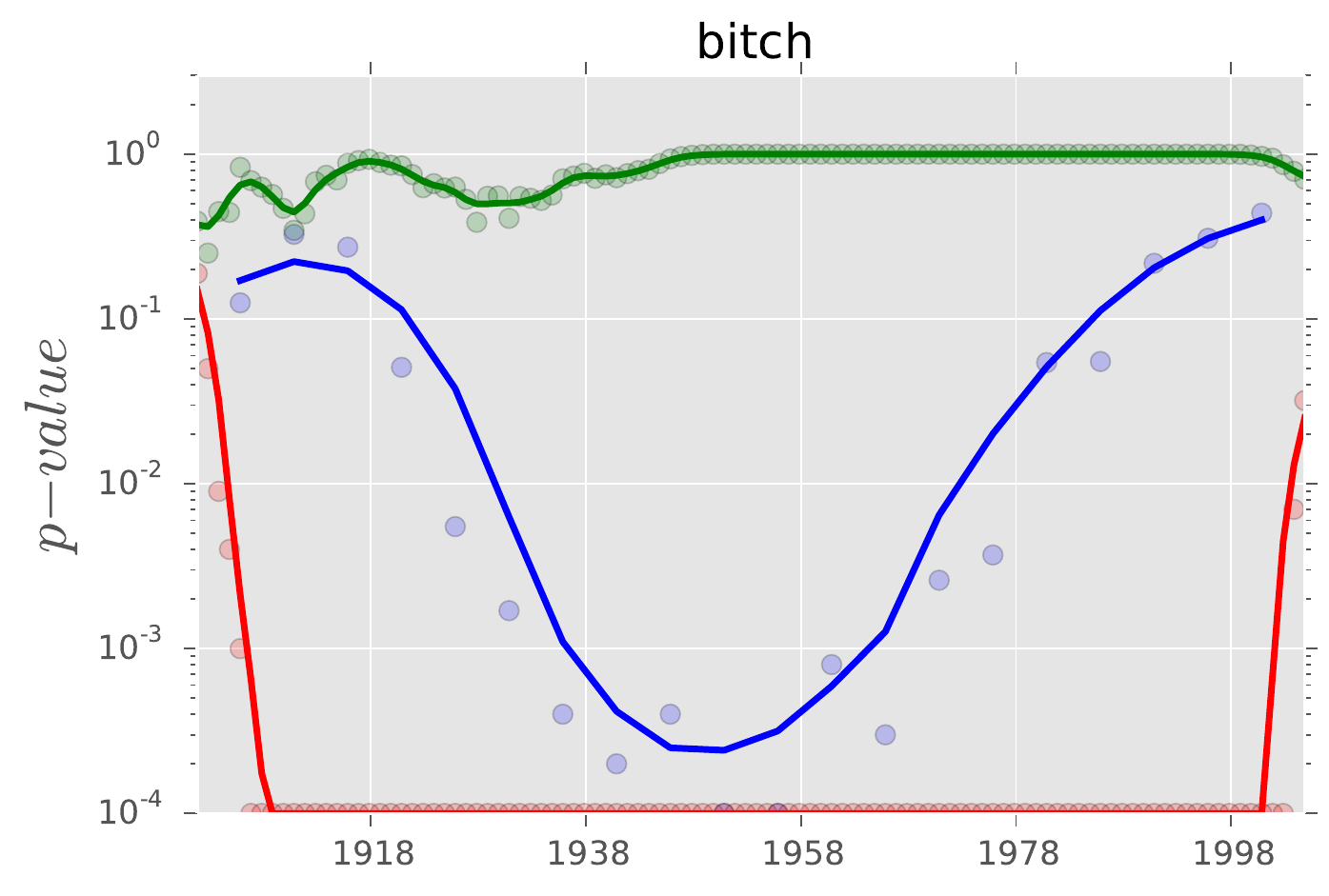} \\
\texttt{sex} & \includegraphics[scale=0.20]{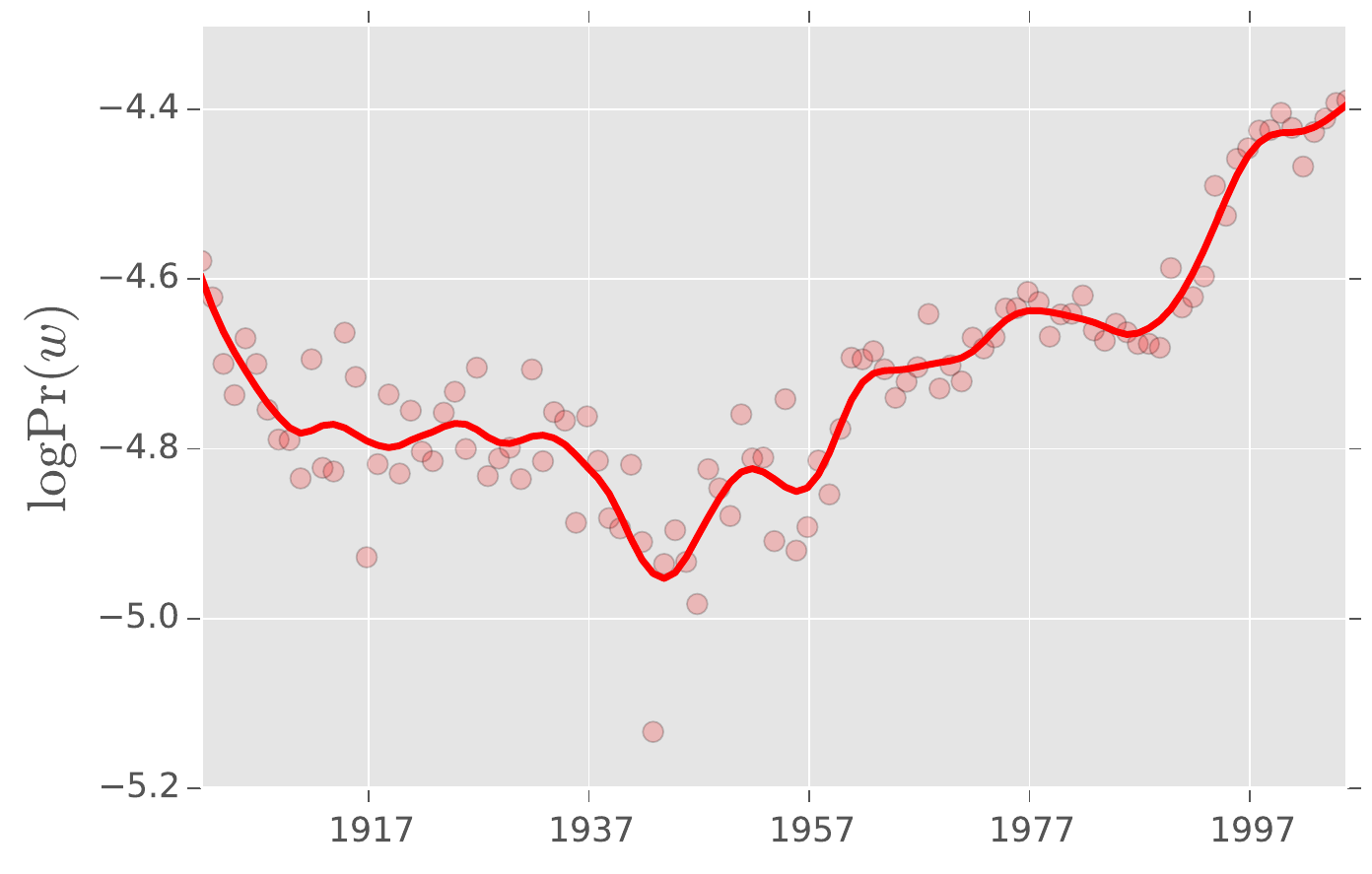} & \includegraphics[scale=0.20]{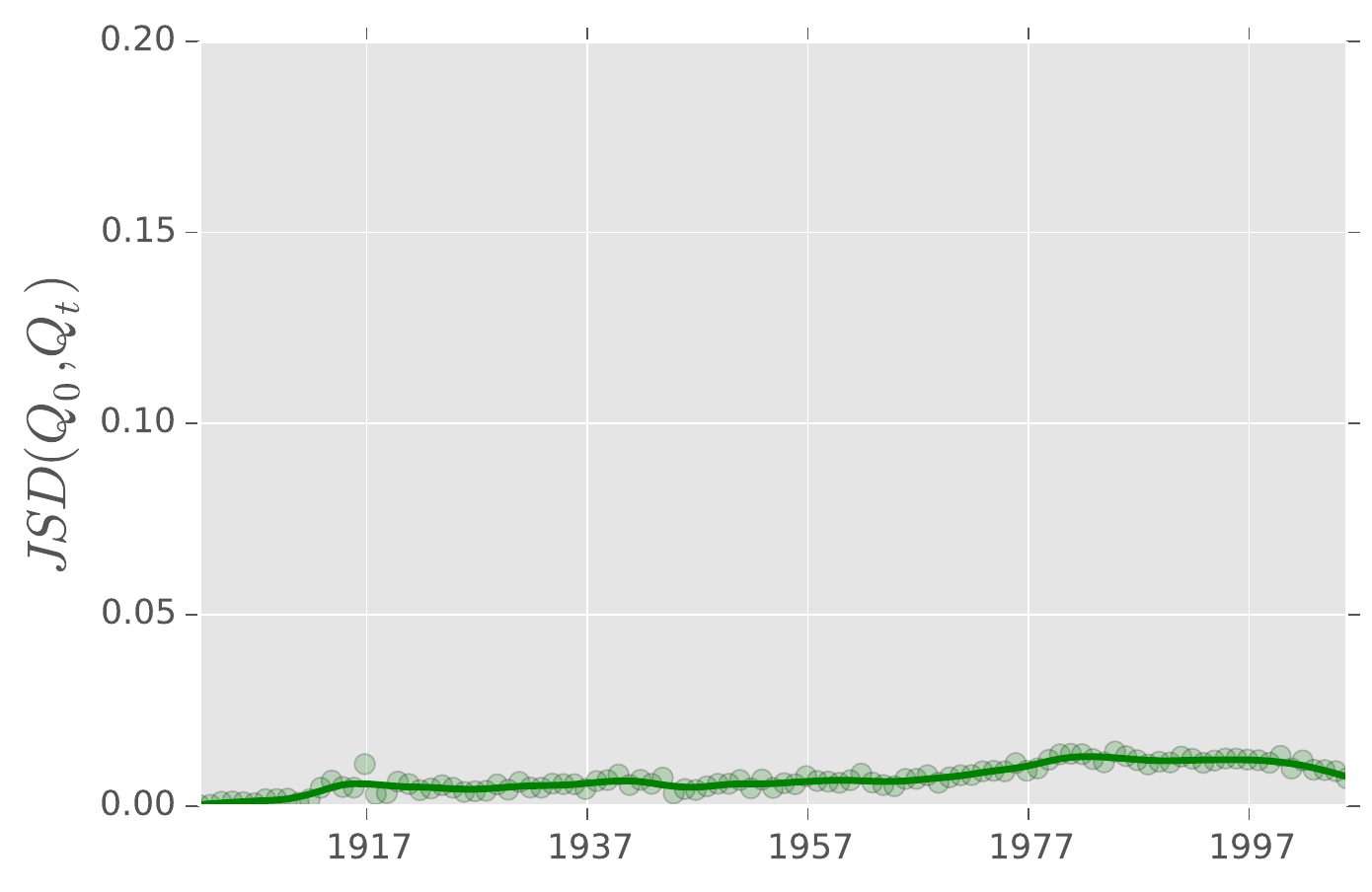}  & \includegraphics[scale=0.20]{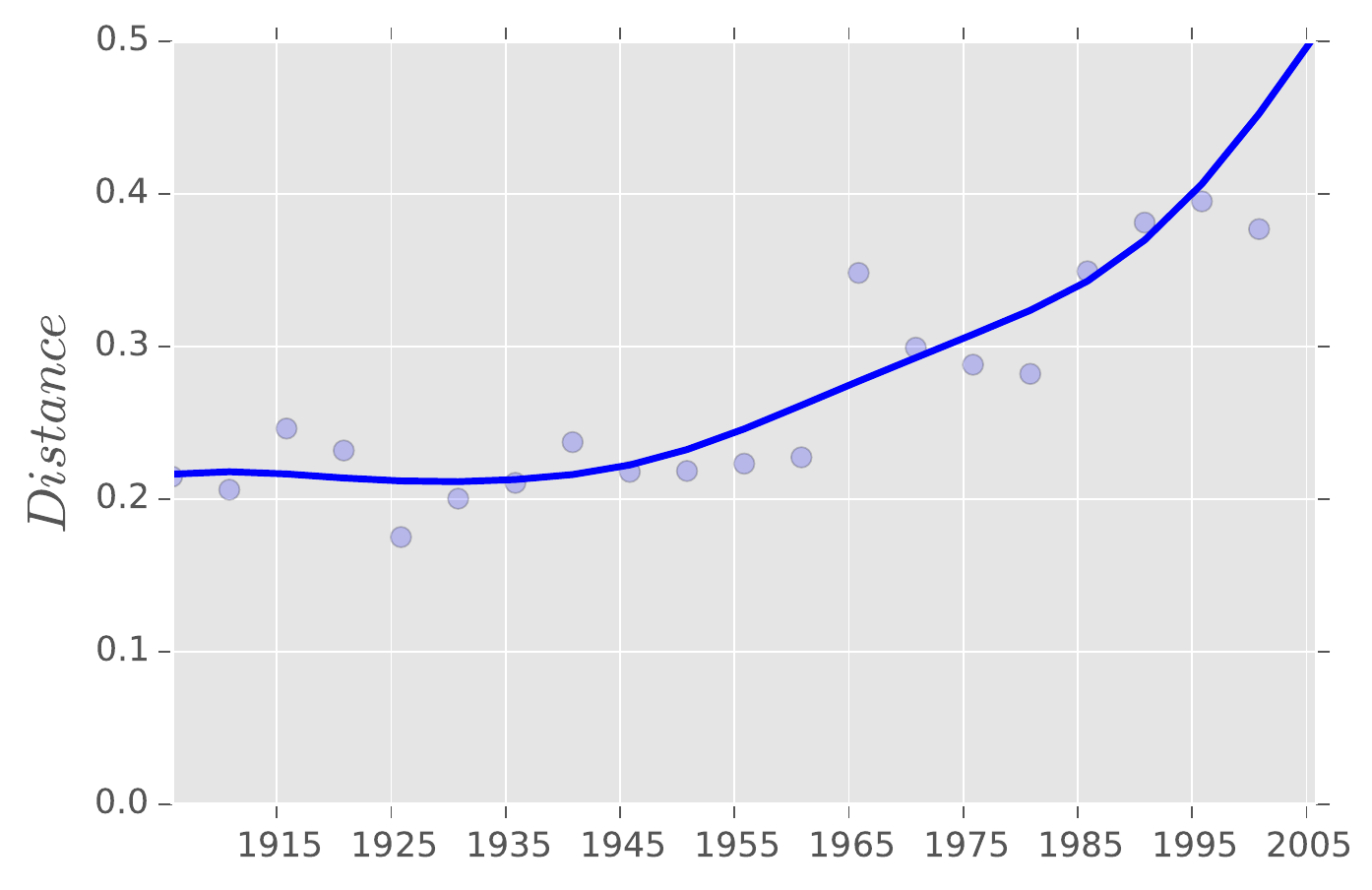}  & \includegraphics[scale=0.20]{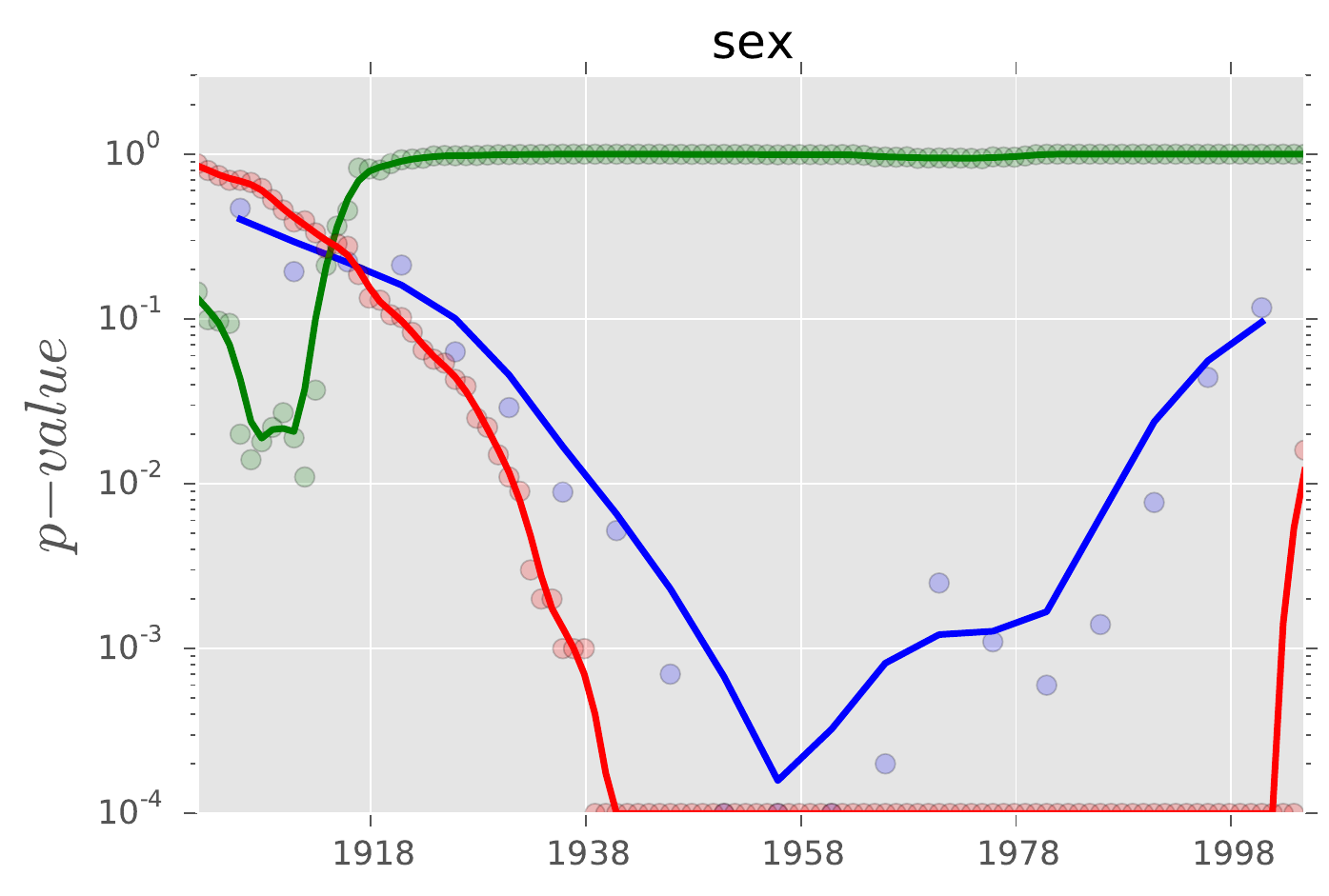} \\
\texttt{her} & \includegraphics[scale=0.20]{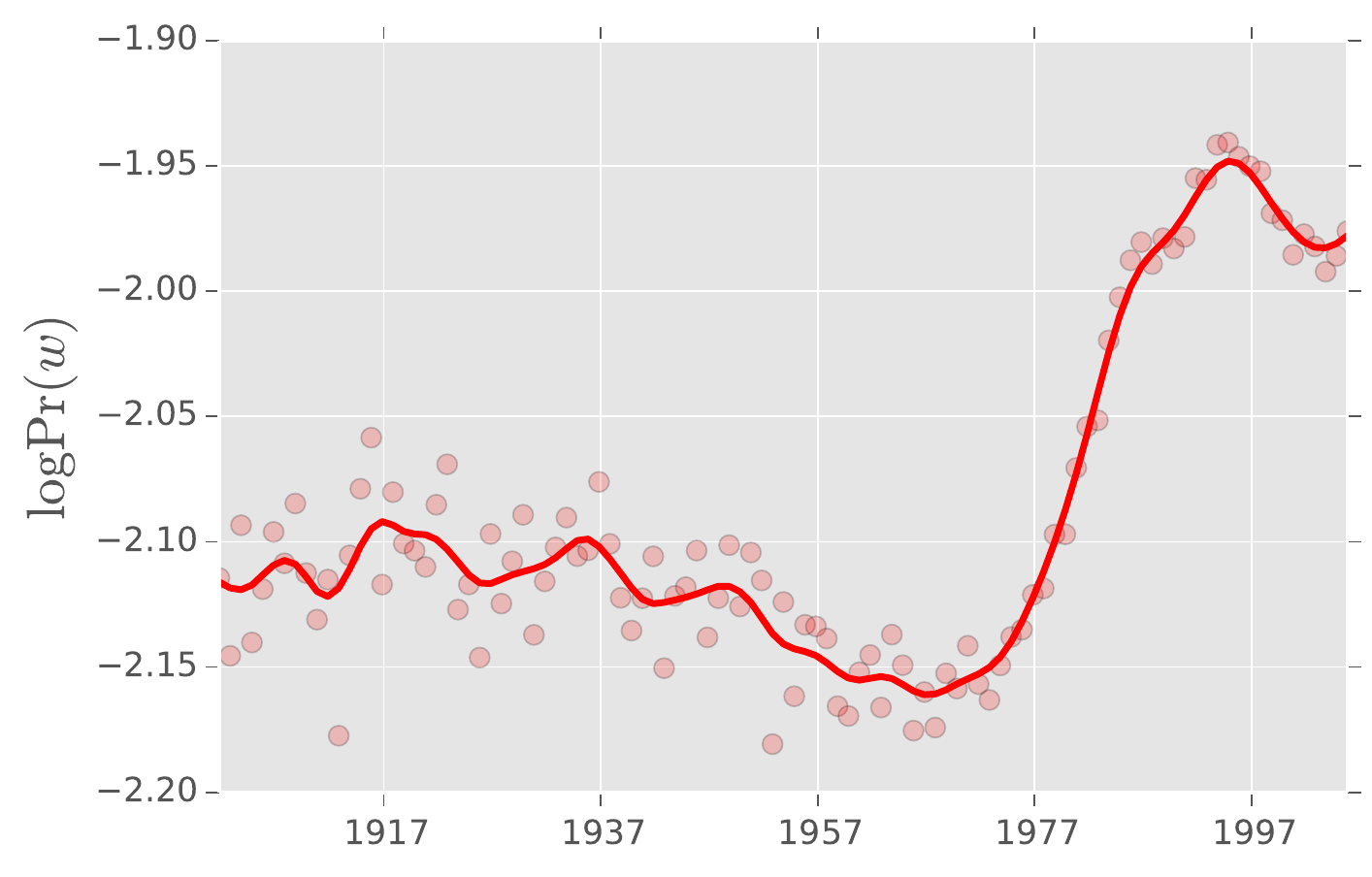} & \includegraphics[scale=0.20]{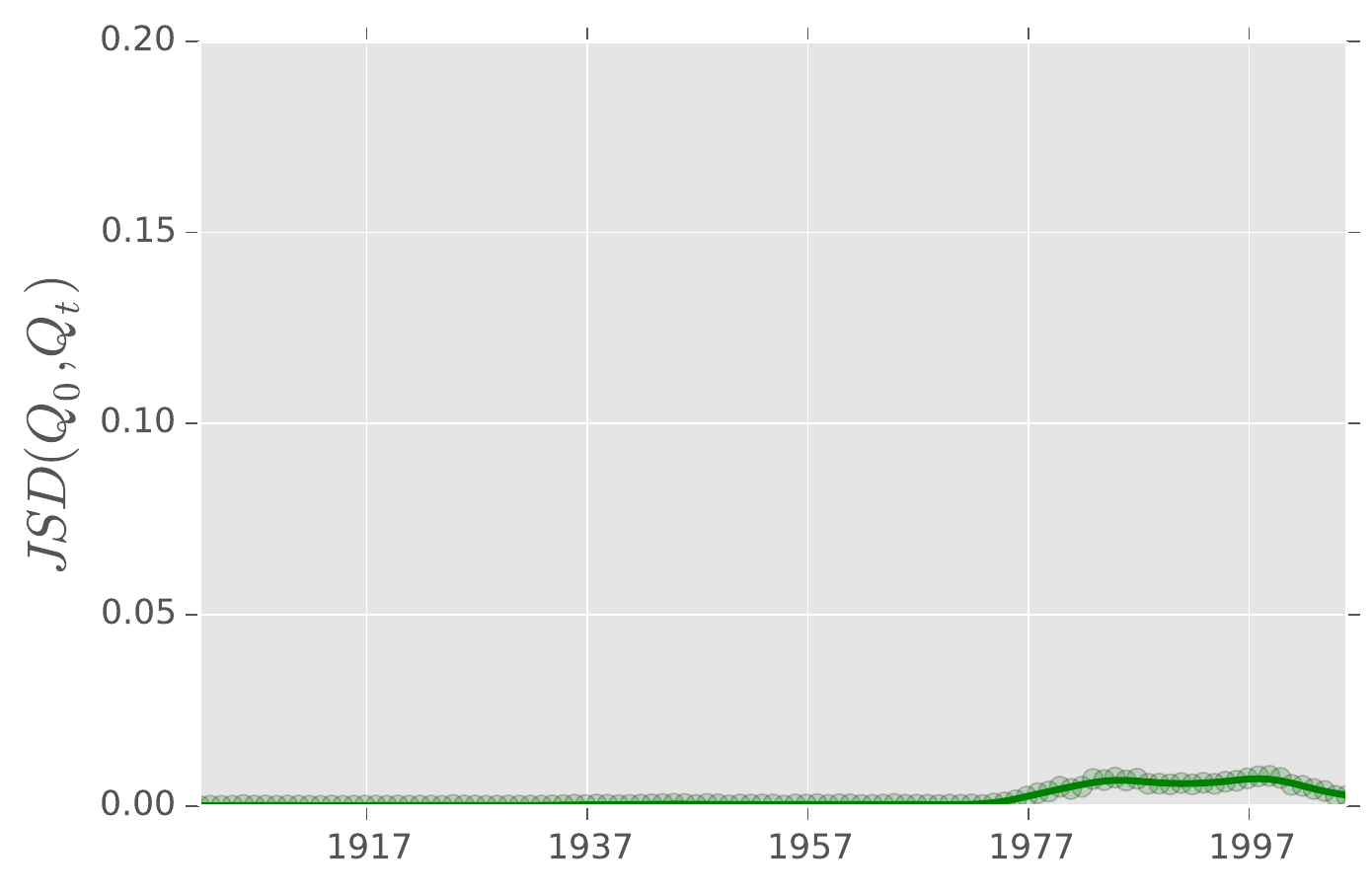}  & \includegraphics[scale=0.20]{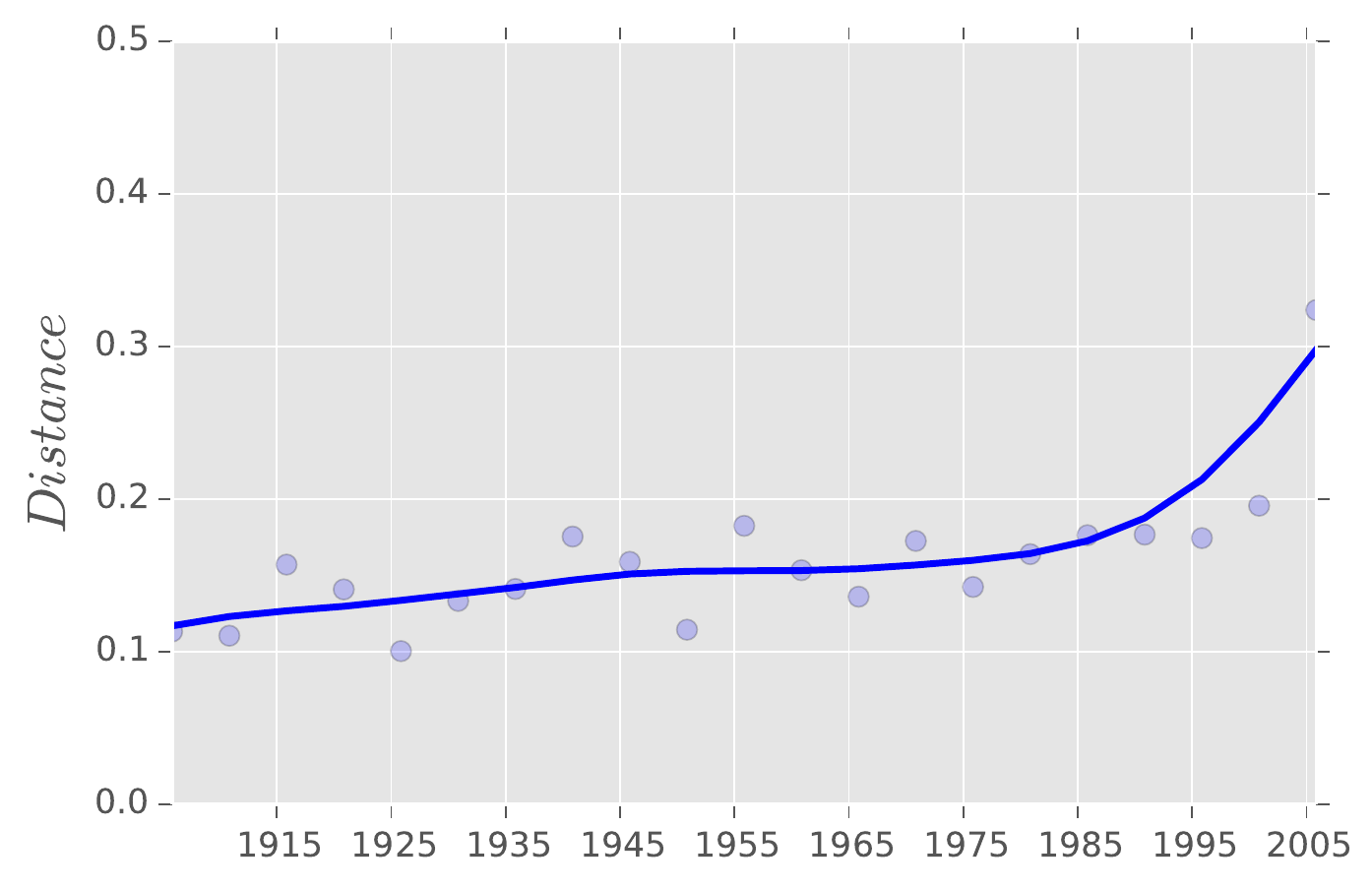}  & \includegraphics[scale=0.20]{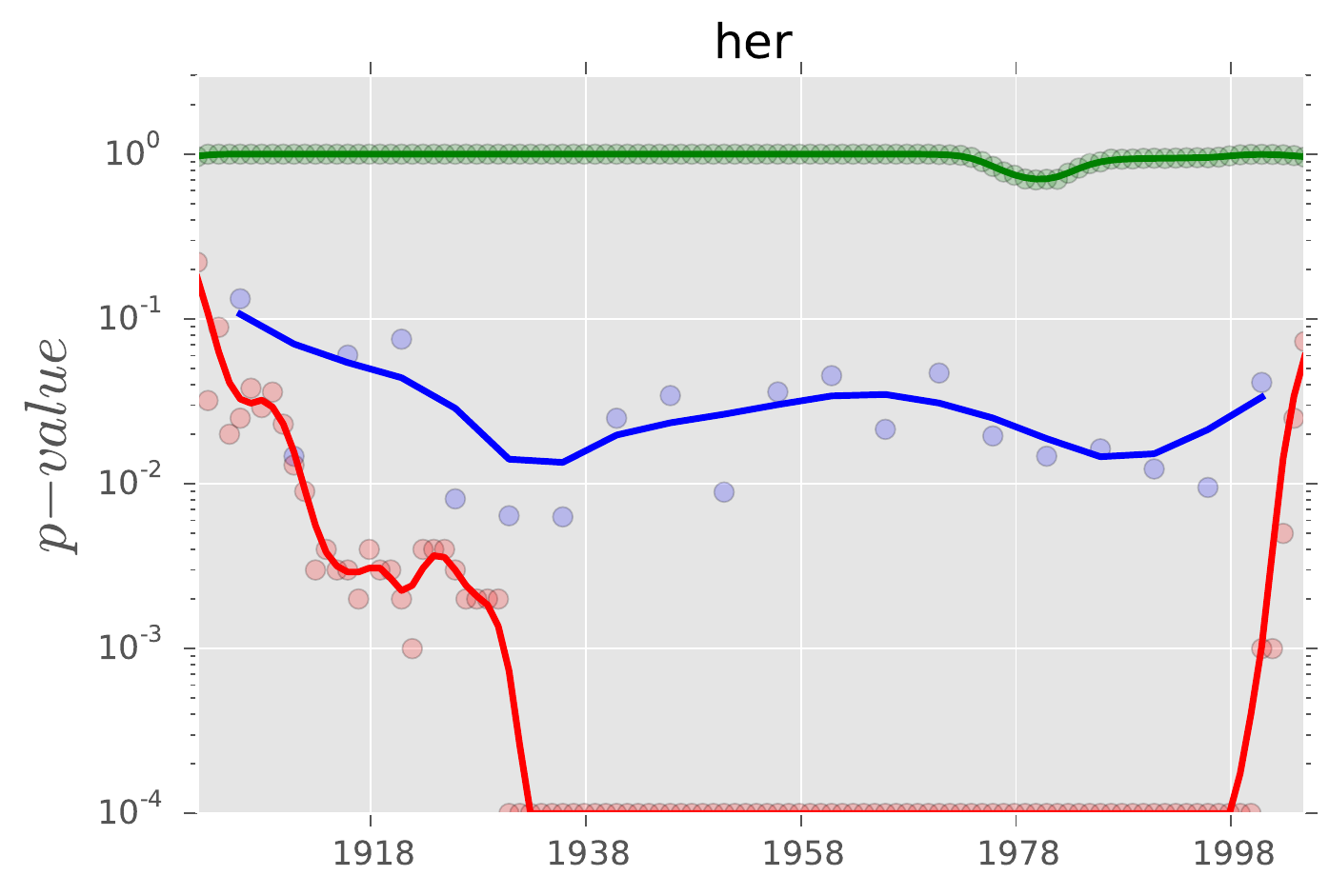} \\
\texttt{desk} & \includegraphics[scale=0.20]{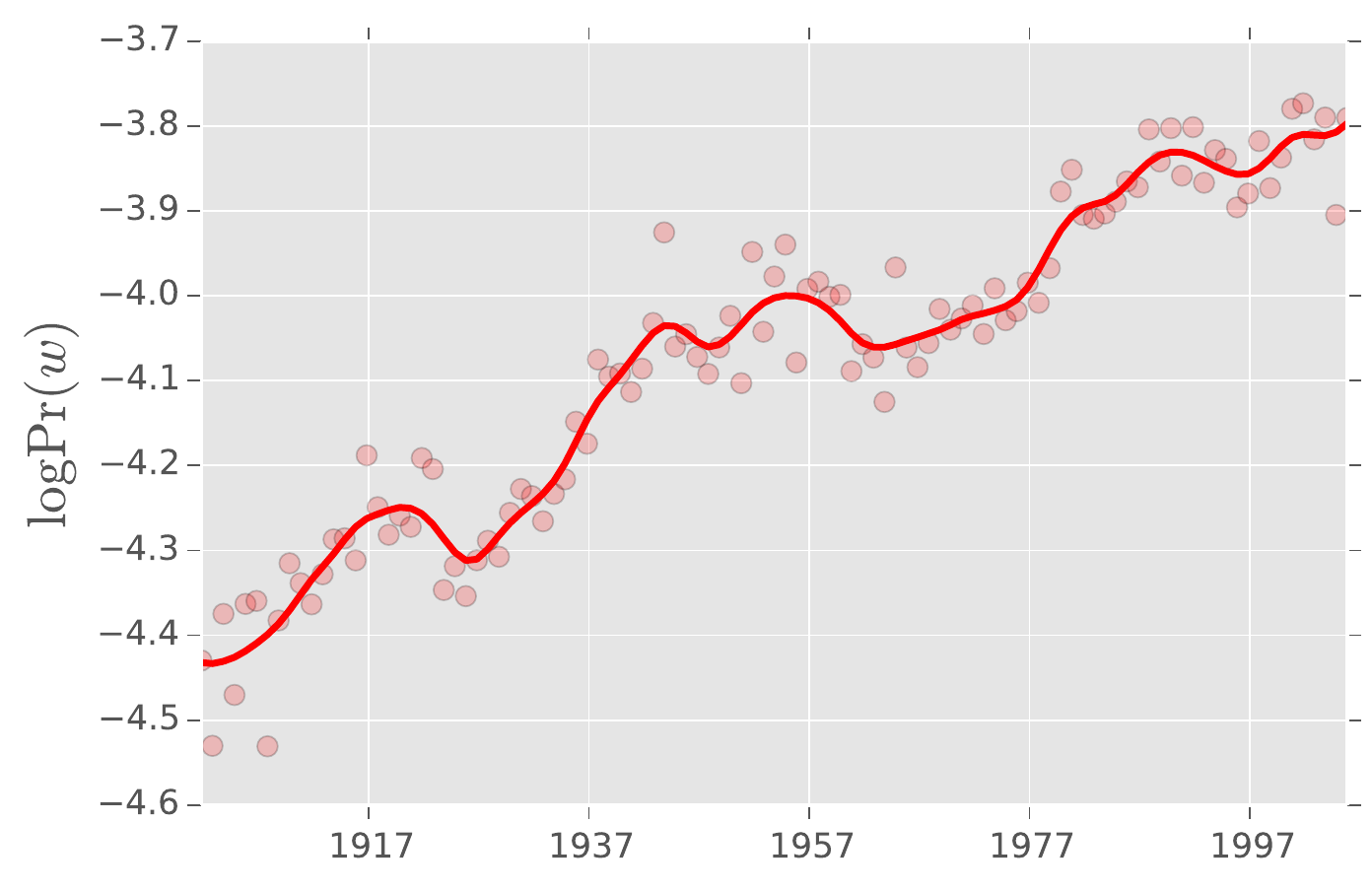} & \includegraphics[scale=0.20]{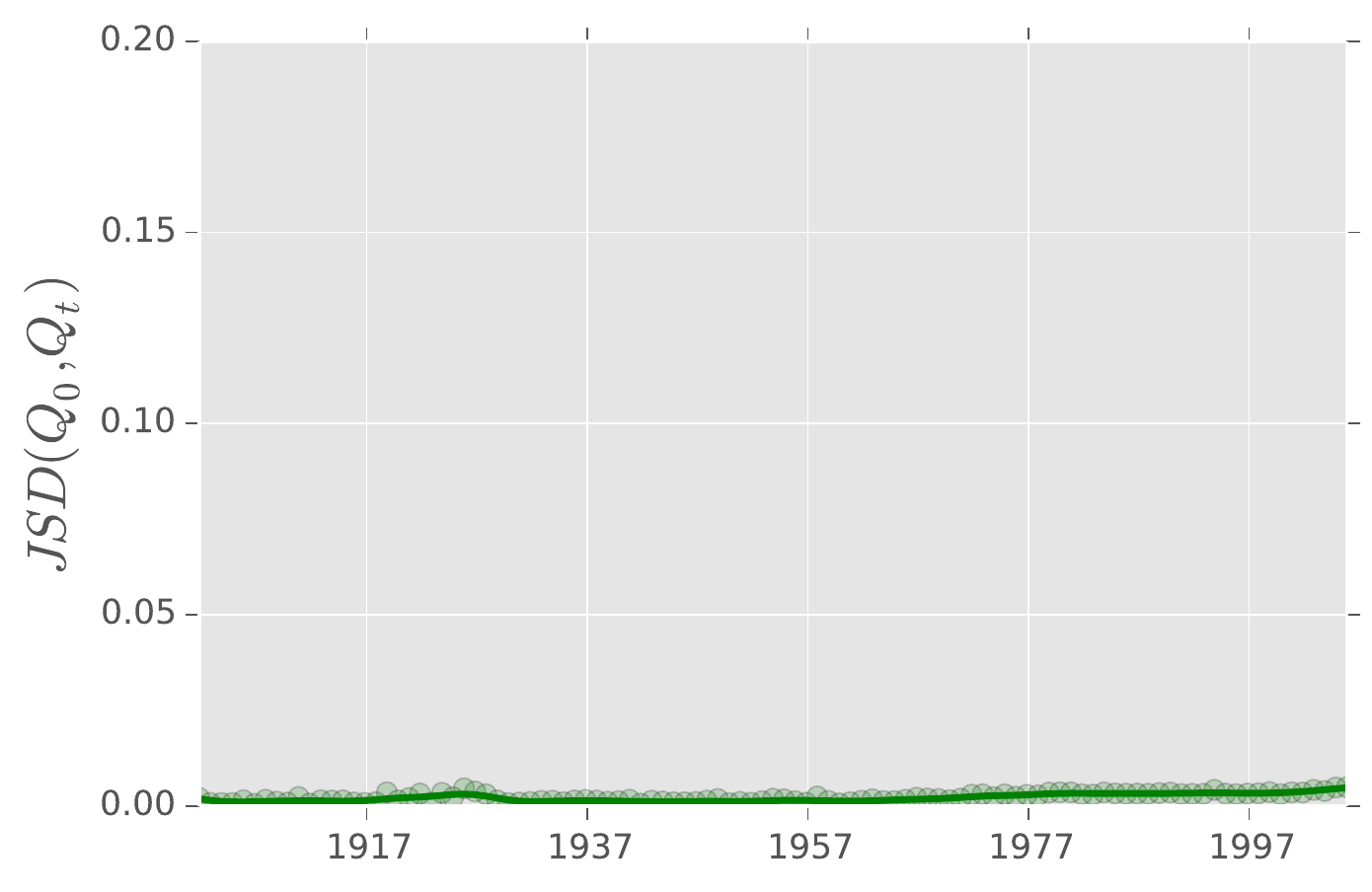}  & \includegraphics[scale=0.20]{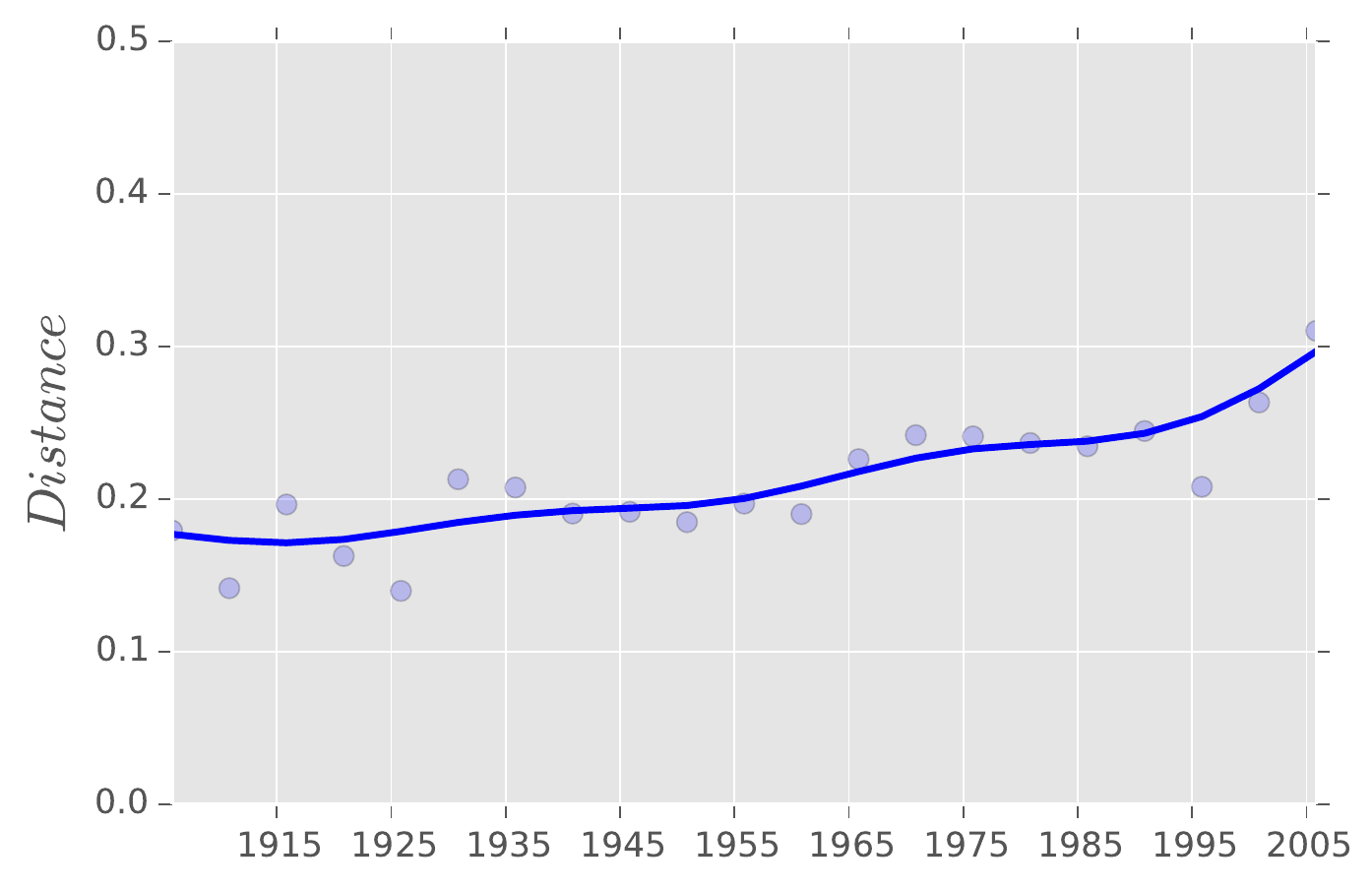}  & \includegraphics[scale=0.20]{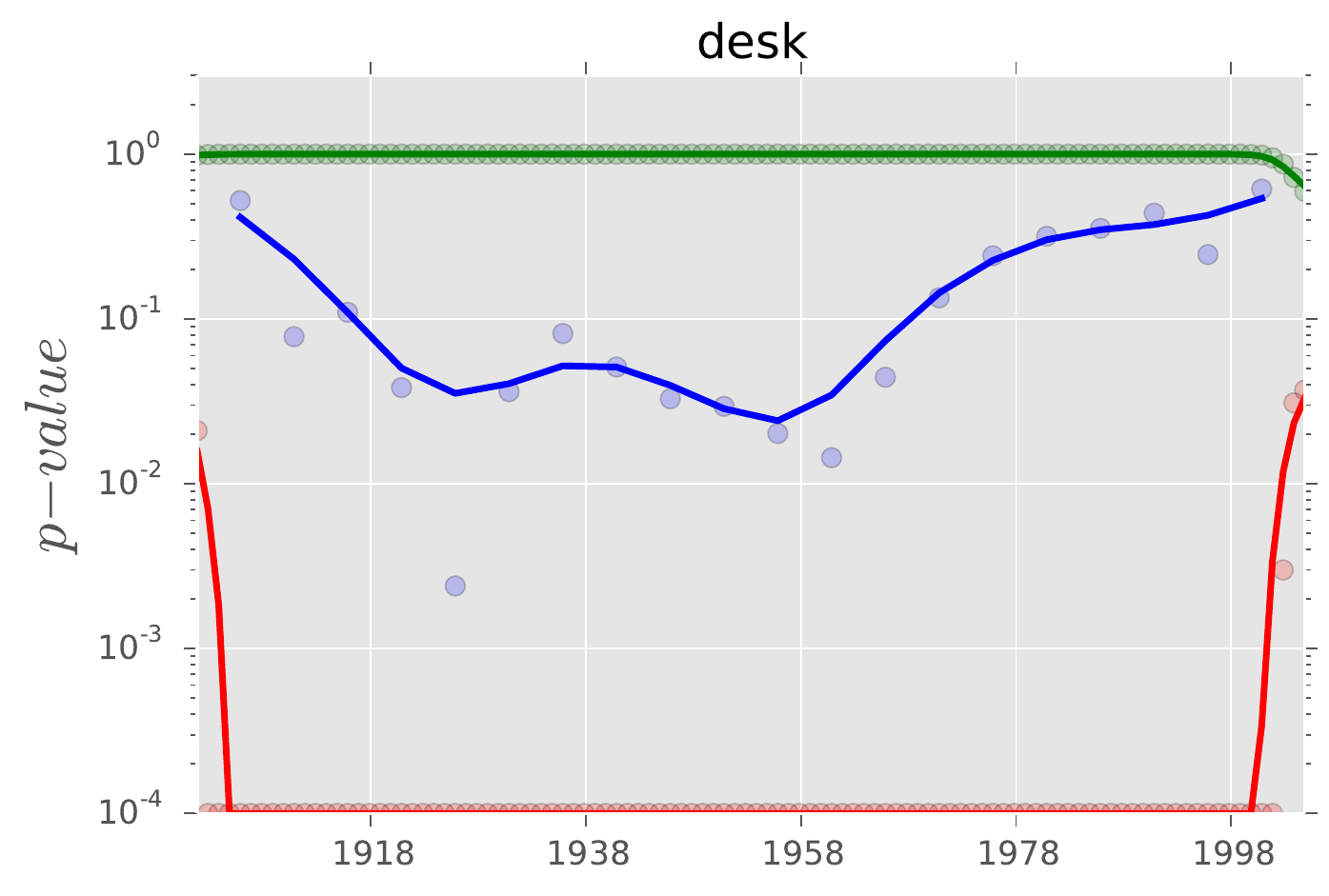} \\
\texttt{apple} & \includegraphics[scale=0.20]{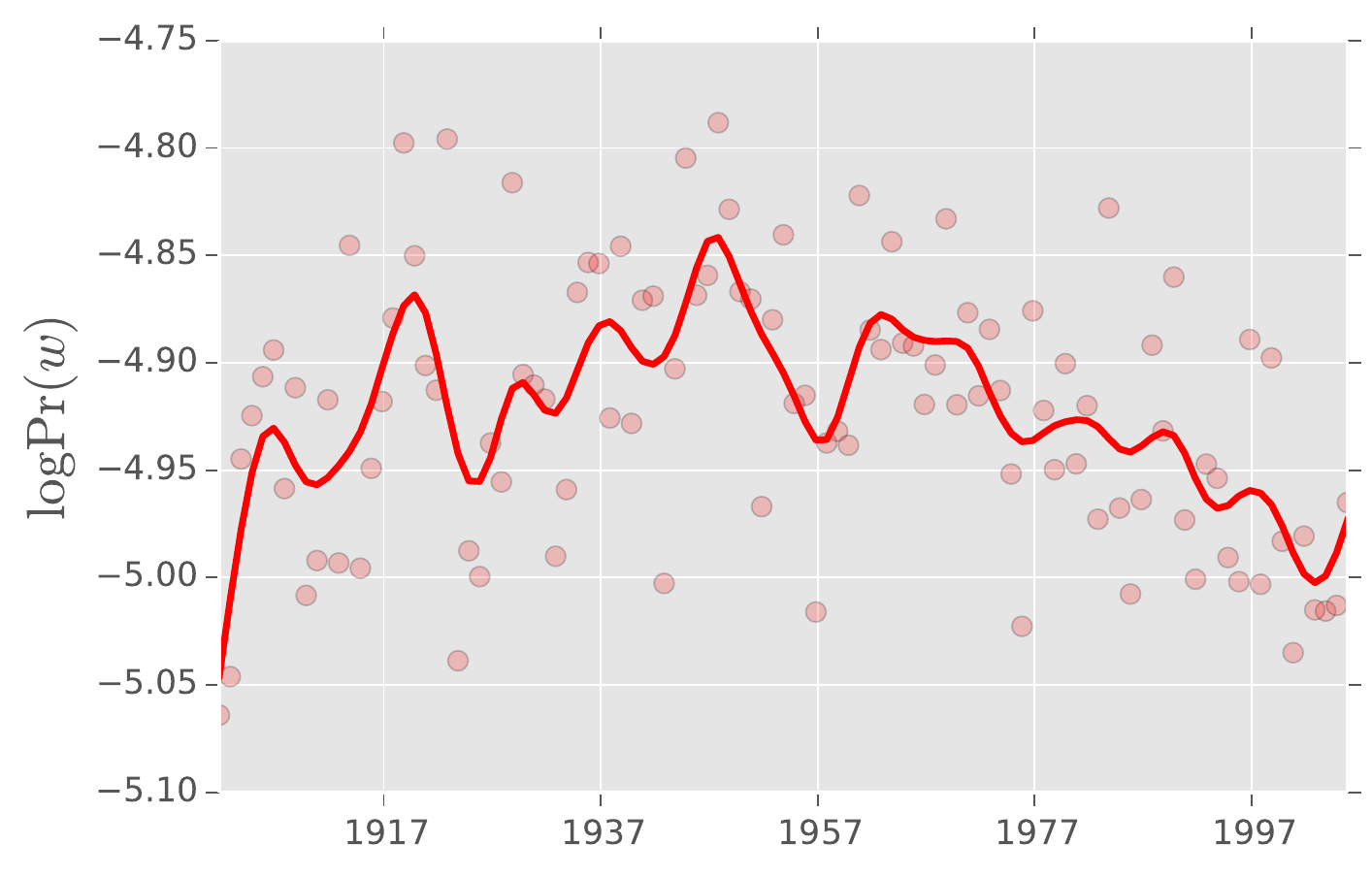} & \includegraphics[scale=0.20]{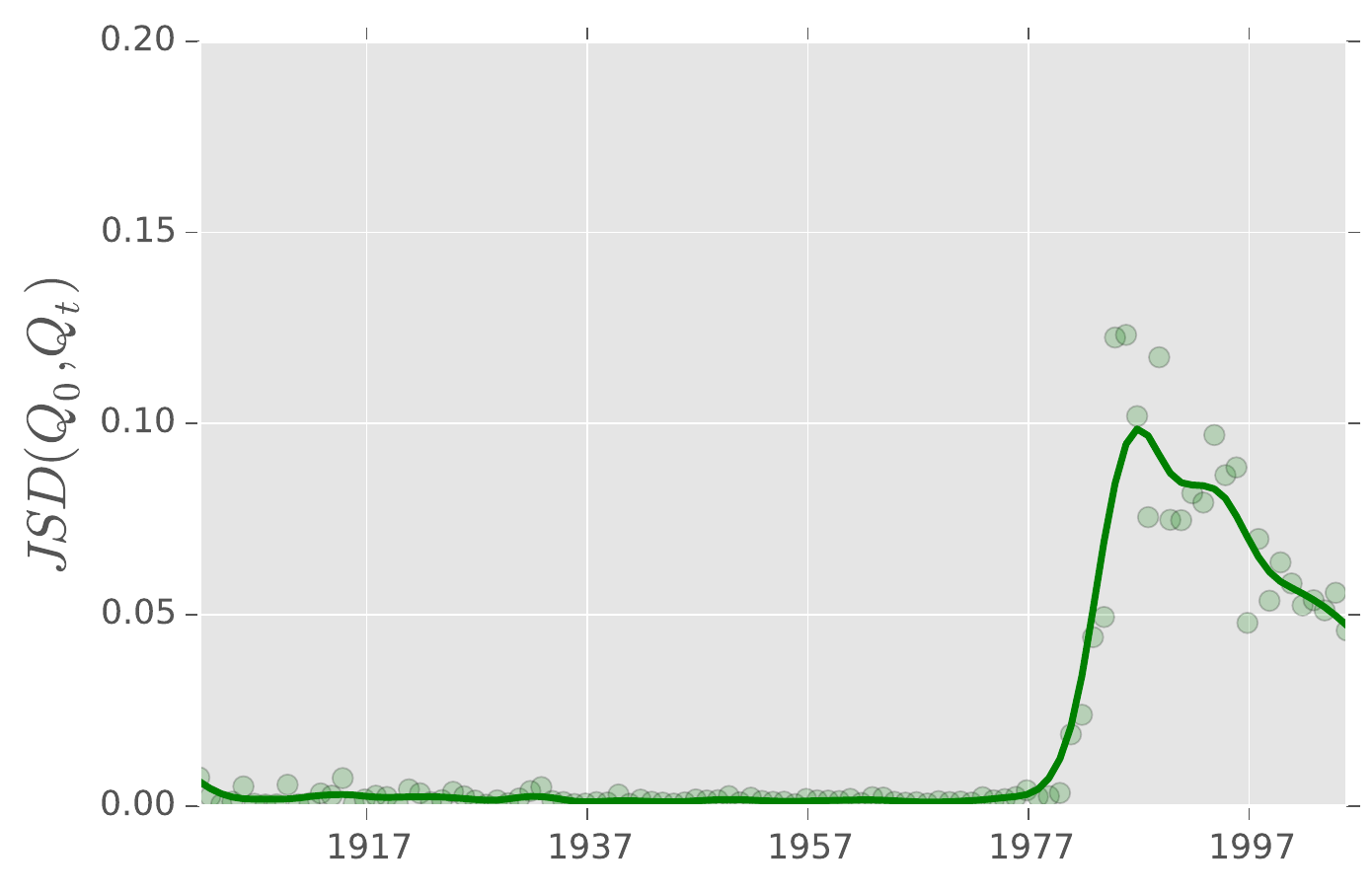}  & \includegraphics[scale=0.20]{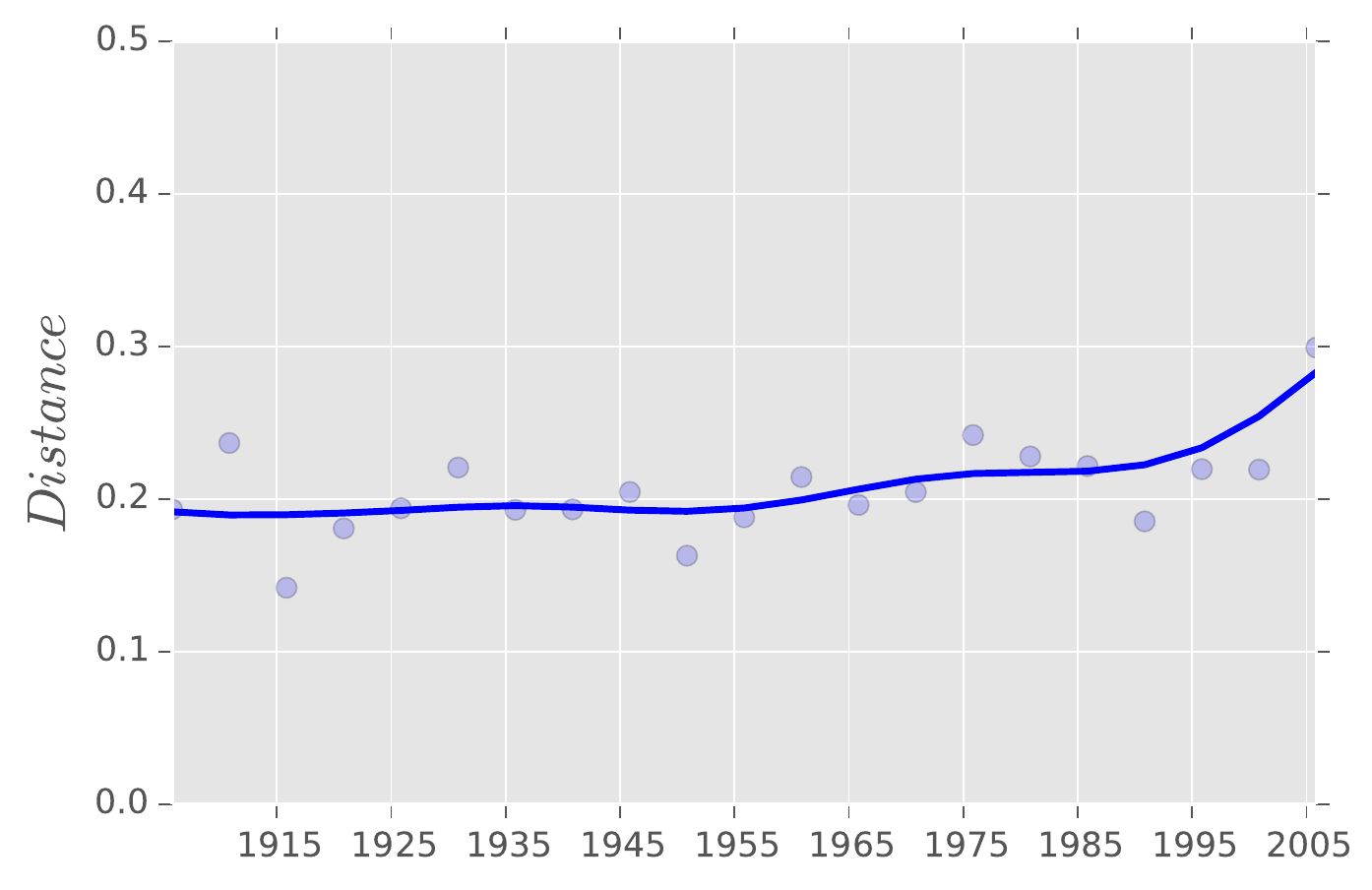}  & \includegraphics[scale=0.20]{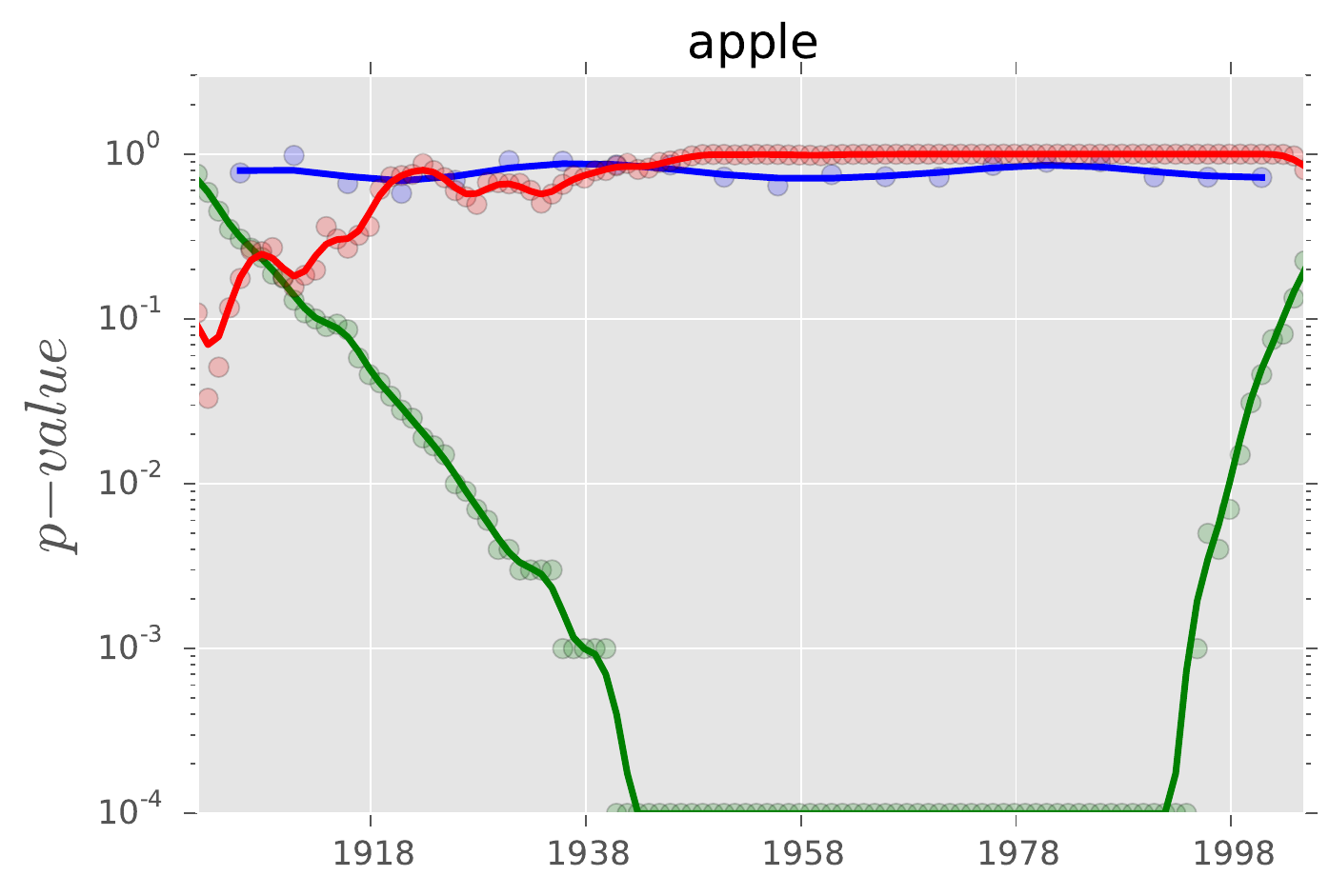} \\
\texttt{diet} & \includegraphics[scale=0.20]{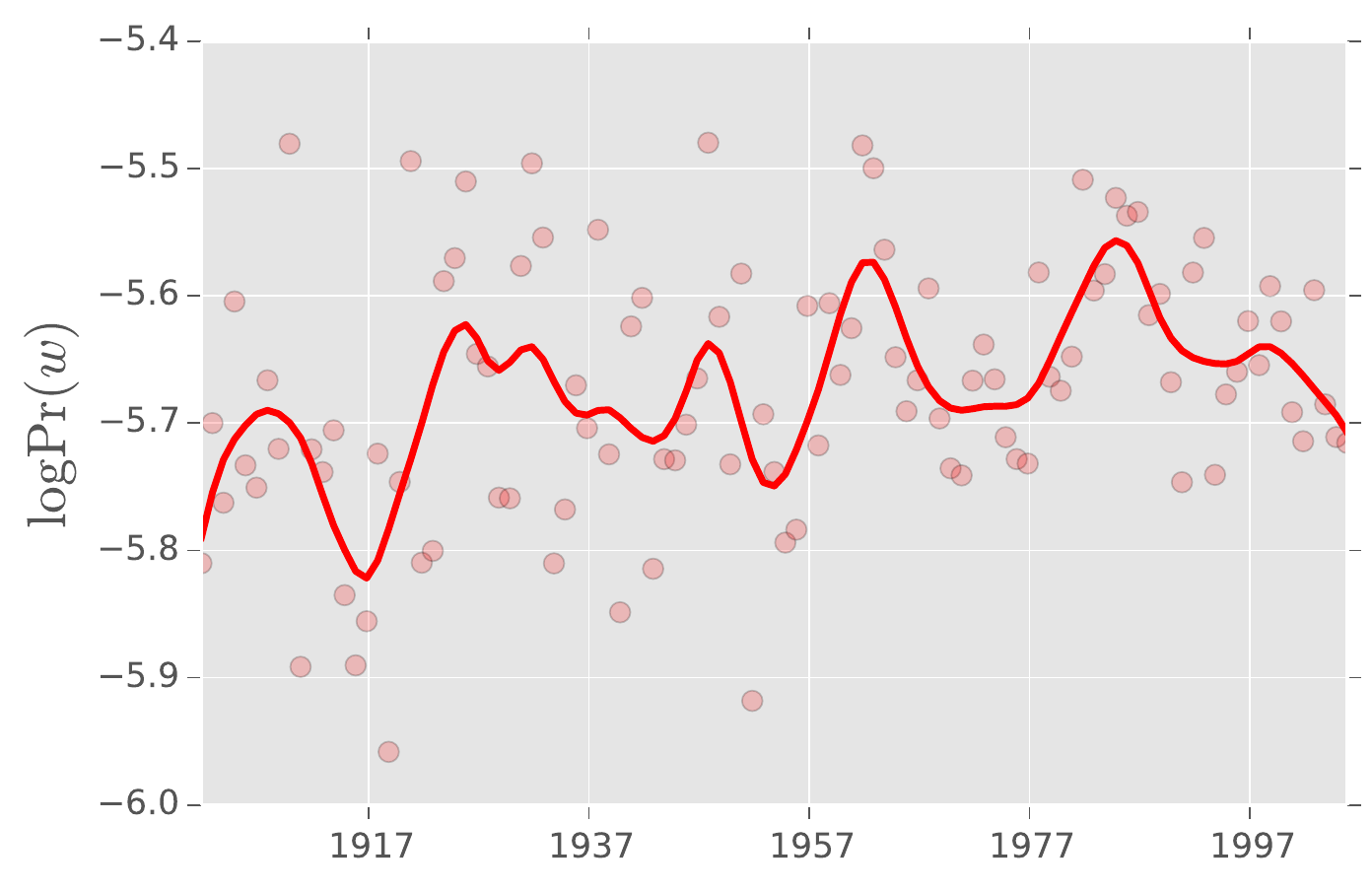} & \includegraphics[scale=0.20]{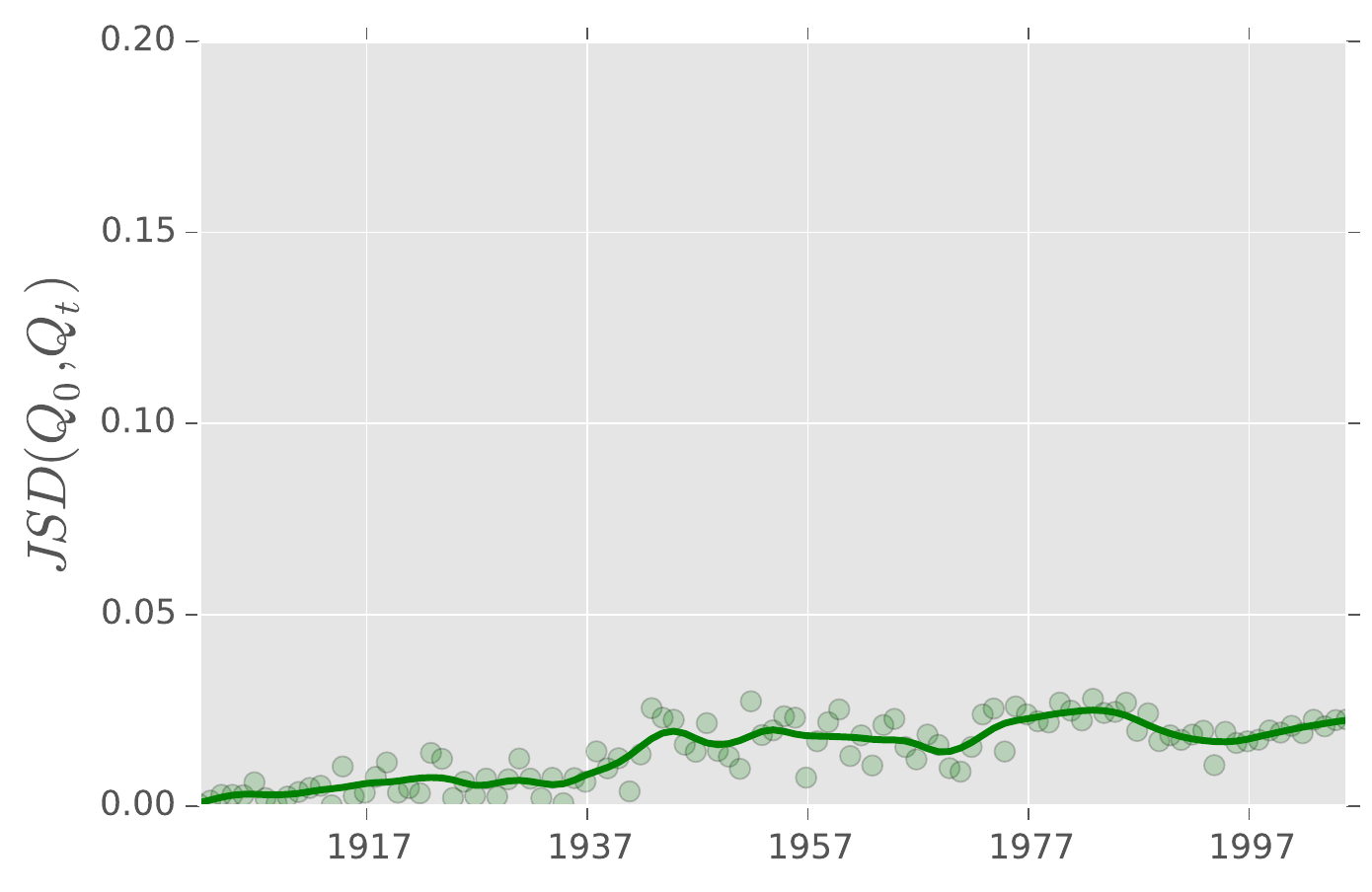}  & \includegraphics[scale=0.20]{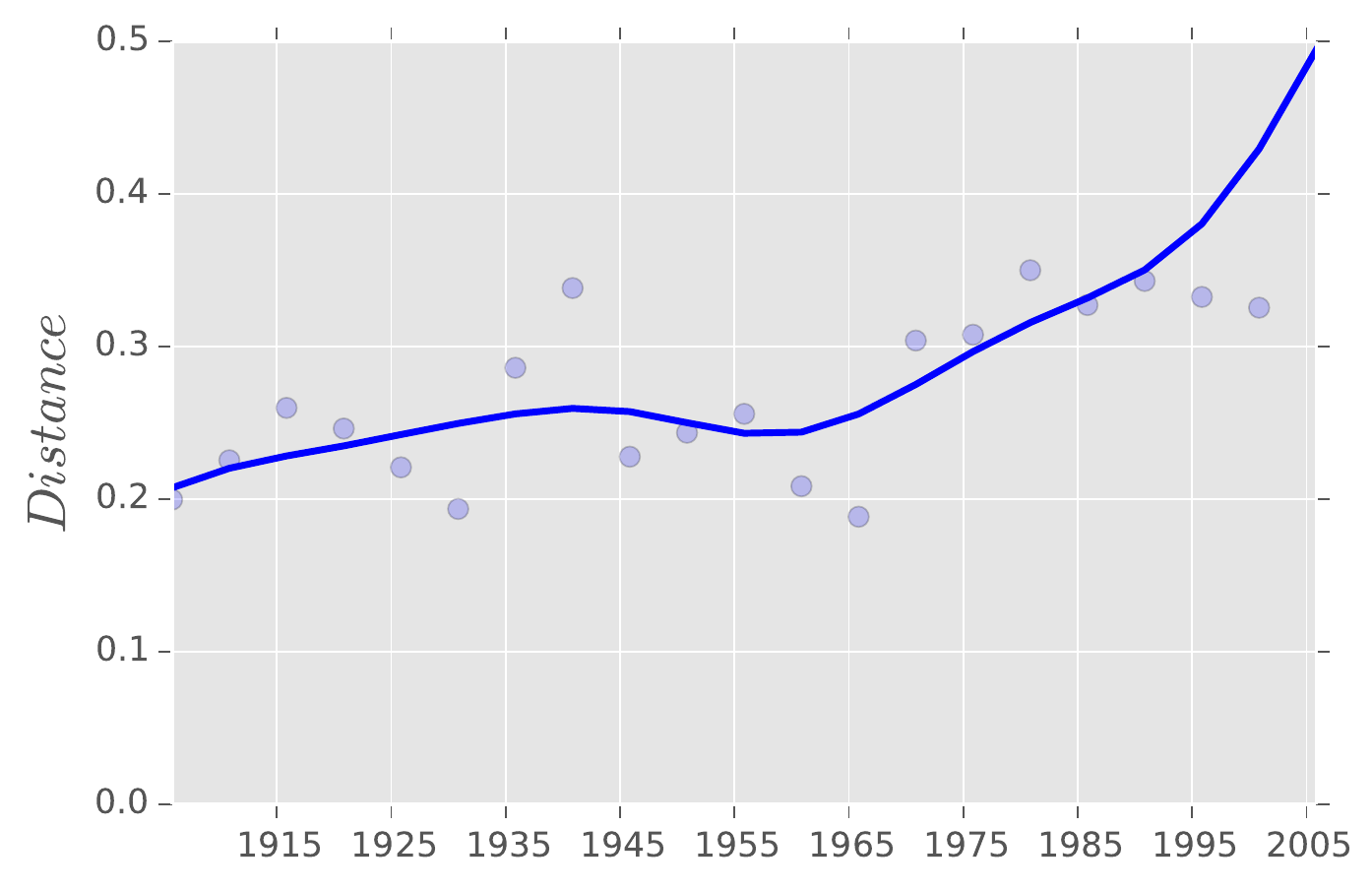}  & \includegraphics[scale=0.20]{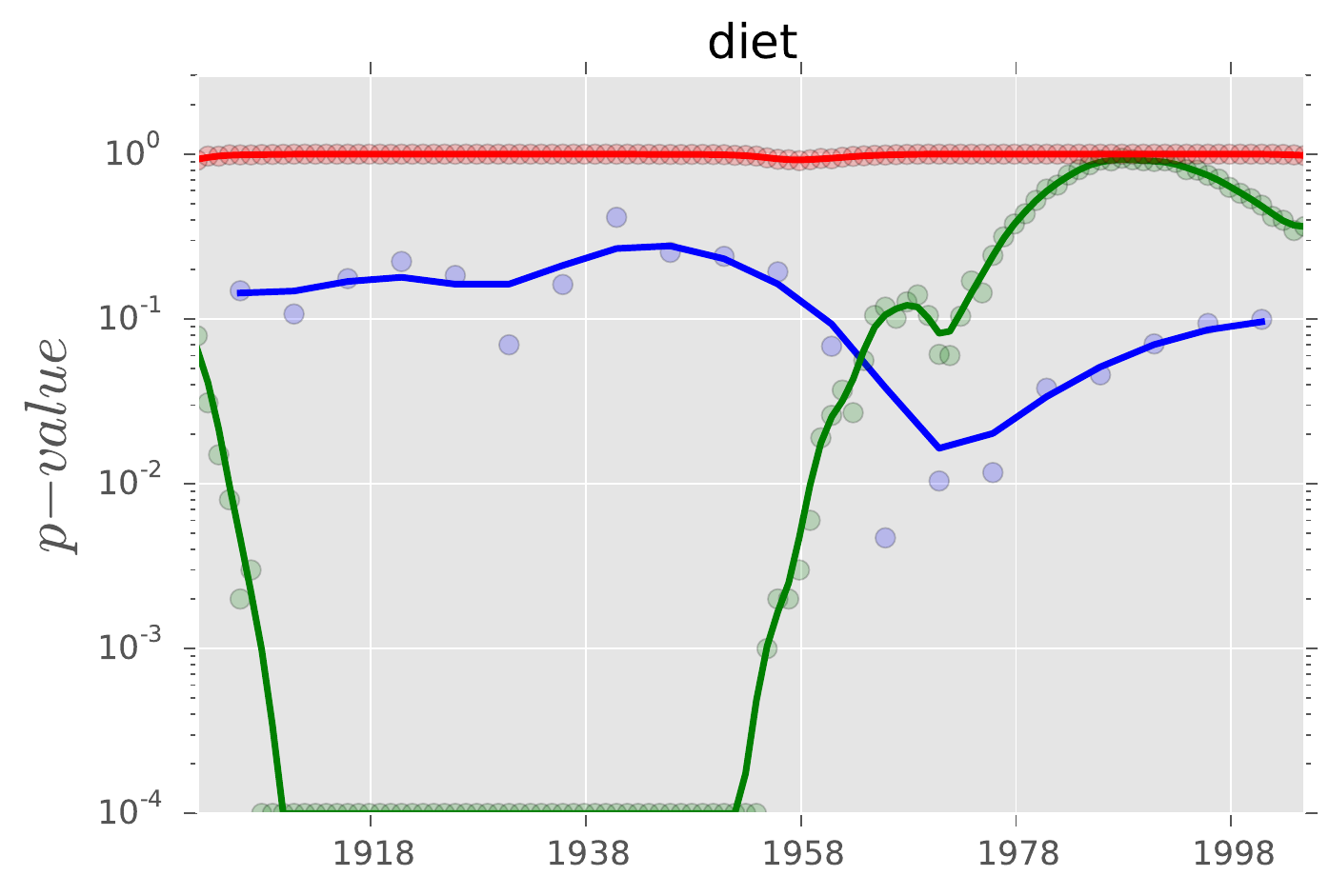} \\
\\
& & {\hwplotA} \freq &  {\hwplotB} \syn & {\hwplotC} \dist \\
\end{tabular}
\caption{Comparison of our different methods of constructing linguistic shift time series on the \ngrams . The first three columns represent time series for a sample of words. The last column shows the \pvalue\ as generated by our point detection algorithm for each method.}
\label{tab:gt}
\end{table*}

\subsection{Time Series Analysis}
As we shall see in Section \ref{sec:synthetic}, our proposed time series construction methods differ in performance.
Here, we use the detected words to study the behavior of our construction methods.

\begin{table*}[t!]
\begin{center}
\begin{tabular}{l|l|cS[table-format = <1.4]|p{55mm}|p{55mm}}

& \textbf{Word} & \textbf{ECP}  & \textbf{\pvalue} & \textbf{Past ngram} & \textbf{Present ngram} \\
\hline
&&&&& \\
& \texttt{recording} & 1990 & 0.0263 & \emph{to be ashamed of recording that} & \emph{recording, photocopying} \\
\parbox[t]{1em}{\multirow{10}{*}{\rotatebox[origin=c]{90}{\dist\ better}}} & \texttt{gay} & 1985 & 0.0001 & \emph{happy and gay} & \emph{gay and lesbians} \\

& \texttt{tape}& 1970 & <0.0001 & \emph{red tape, tape from her mouth}  & \emph{a copy of the tape} \\

& \texttt{checking} & 1970 & 0.0002 & \emph{then checking himself} & \emph{checking him out} \\

& \texttt{diet} & 1970 & 0.0104 & \emph{diet of bread and butter} & \emph{go on a diet} \\

& \texttt{sex} & 1965 & 0.0002 & \emph{and of the fair sex} & \emph{have sex with} \\

& \texttt{bitch} & 1955 & 0.0001 & \emph{nicest black bitch} (Female dog) & \emph{bitch} (Slang) \\

& \texttt{plastic} & 1950 & 0.0005 & \emph{of plastic possibilities} & \emph{put in a plastic} \\

& \texttt{transmitted} & 1950 & 0.0002 & \emph{had been transmitted to him, transmitted from age to age} & \emph{transmitted in electronic form} \\

& \texttt{peck} & 1935 & 0.0004 & \emph{brewed a peck} & \emph{a peck on the cheek} \\

& \texttt{honey} & 1930 & 0.01 & \emph{land of milk and honey} & \emph{Oh honey!} \\

\multicolumn{6}{c}{} \\
& & &  & \textbf{Past POS} & \textbf{Present POS} \\ \hline
&&&&& \\

\parbox[t]{1em}{\multirow{8}{*}{\rotatebox[origin=c]{90}{\syn \hspace{1pt} better}}}

& \texttt{hug} & 2002 & <0.001 & Verb (\emph{hug a child}) & Noun (\emph{a free hug}) \\

& \texttt{windows} & 1992 & <0.001  & Noun (\emph{doors and windows of a house}) & Proper Noun (\emph{Microsoft Windows}) \\

& \texttt{bush} & 1989 & <0.001 & Noun (\emph{bush and a shrub}) & Proper Noun (\emph{George Bush})\\

& \texttt{apple} & 1984 & <0.001 & Noun (\emph{apple, orange, grapes}) & Proper Noun (\emph{Apple computer})\\
& \texttt{sink} & 1972 & <0.001 & Verb (\emph{sink a ship}) & Noun (\emph{a kitchen sink}) \\

& \texttt{click} & 1952 & <0.001 & Noun (\emph{click of a latch}) & Verb (\emph{click a picture}) \\

& \texttt{handle} & 1951 & <0.001 & Noun (\emph{handle of a door}) & Verb (\emph{he can handle it}) \\

\end{tabular}
\end{center}
\caption{Estimated change point (ECP) as detected by our approach for a sample of words on \ngrams.\dist\ method is better on some words (which \syn\ did not detect as statistically significant eg. sex, transmitted, bitch, tape, peck) while \syn\ method is better on others (which \dist\ failed to detect as statistically significant eg. apple, windows, bush)}
\label{tab:ngrams_table}
\end{table*}

\begin{table*}[t!]
\begin{center}
\begin{tabular}{l|l|S[table-format = <1.4]|r|p{55mm}|p{55mm}}

&\textbf{Word} & \textbf{\pvalue} & \textbf{ECP} & \textbf{Past Usage} & \textbf{Present Usage} \\
\hline
& & &  & & \\

\parbox[t]{1em}{\multirow{7}{*}{\rotatebox[origin=c]{90}{\small \textbf{Amazon Reviews}}}}
& \texttt{instant} & 0.016 & 2010 & \emph{instant hit, instant dislike} & \emph{instant download}\\
& \texttt{twilight} & 0.022 & 2009 & \emph{twilight as in dusk} & \emph{Twilight} (The movie)\\
& \texttt{rays} & 0.001 & 2008 & \emph{x-rays} & \emph{blu-rays}\\
& \texttt{streaming} & 0.002 & 2008 & \emph{sunlight streaming} & \emph{streaming video}\\
& \texttt{ray} & 0.002 & 2006 & \emph{ray of sunshine} & \emph{Blu-ray}\\
& \texttt{delivery} & 0.002 & 2006 & \emph{delivery of dialogue} & \emph{timely delivery of products} \\
& \texttt{combo} & 0.002 & 2006 & \emph{combo of plots} & \emph{combo DVD pack} \\

\multicolumn{6}{c}{} \\

\parbox[t]{1em}{\multirow{7}{*}{\rotatebox[origin=c]{90}{\textbf{Tweets}}}}
& \texttt{candy} & <0.001 & Apr $2013$ & \emph{candy sweets} & \emph{Candy Crush} (The game) \\
& \texttt{rally} & <0.001 &  Mar $2013$ & \emph{political rally} & \emph{rally of soldiers} (Immortalis game)\\
& \texttt{snap} & <0.001 & Dec $2012$ & \emph{snap a picture} & \emph{snap chat}\\
& \texttt{mystery} & <0.001 & Dec $2012$ & \emph{mystery books}  & \emph{Mystery Manor} (The game)\\
& \texttt{stats} & <0.001 & Nov $2012$ &  \emph{sport statistics} & \emph{follower statistics}\\
& \texttt{sandy} & 0.03 & Sep $2012$ & \emph{sandy beaches} & \emph{Hurricane Sandy}\\
& \texttt{shades} & <0.001 & Jun $2012$ & \emph{color shade, shaded glasses} & \emph{$50$ shades of grey} (The Book)\\
\end{tabular}
\end{center}
\caption{Sample of words detected by our \dist\ method on Amazon Reviews and Tweets.}
\label{tab:internet_data}
\end{table*}

Table \ref{tab:gt} shows the time series constructed for a sample of words with their corresponding \pvalue\ time series, displayed in the last column.
A dip in the \pvalue\ is indicative of a shift in the word usage.
The first three words, \texttt{transmitted}, \texttt{bitch}, and \texttt{sex}, are detected by both the \freq\ and \dist\ methods.
Table \ref{tab:ngrams_table} shows the previous and current senses of these words demonstrating the changes in usage they have gone through.

Observe that words like \texttt{her} and \texttt{desk} did not change, however, the \freq\ method detects a change.
The sharp increase of the word \texttt{her} in frequency around the 1960's could be attributed to the concurrent rise and popularity of the feminist movement.
Sudden temporary popularity of specific social and political events could lead the \freq\ method to produce many false positives.
These results confirm our intuition we illustrated in Figure \ref{fig:crownjewel}.
While frequency analysis (like Google Trends) is an extremely useful tool to visualize trends, it is not very well suited for the task of detecting linguistic shift.

The last two rows in Table \ref{tab:gt} display two words (\texttt{apple} and \texttt{diet}) that \syn\ method detected.
The word \texttt{apple}\ was detected uniquely by the \syn\ method as its most frequent part of speech tag changed significantly from ``Noun" to ``Proper Noun".
While both \syn\ and \dist\ methods indicate the change in meaning of the word \texttt{diet}, it is only the \dist\ method that detects the right point of change (as shown in Table \ref{tab:ngrams_table}).
The \syn\ method is indicative of having low false positive rate, but suffers from a high false negative rate, given that only two words in the table were detected. Furthermore, observe that \syn\ method relies on good linguistic taggers. However, linguistic taggers require annotated data sets and also do not work well across domains.

We find that the \dist\ method offers a good balance between false positives and false negatives, while requiring no linguistic resources of any sort.
Having analyzed the words detected by different time series we turn our attention to the analysis of estimated changepoints.

\subsection{Historical Analysis}
We have demonstrated that our methods are able to detect words that shifted in meaning.
We seek to identify the inflection points in time where the new senses are introduced.
Moreover, we are interested in understanding how the new acquired senses differ from the previous ones.

Table \ref{tab:ngrams_table} shows sample words that are detected by \syn\ and \dist\ methods.
The first set represents words which the \dist\ method detected (\dist\ better) while the second set shows sample words which \syn\ method detected (\syn\ better).

Our \dist\ method estimates that the word \texttt{tape} changed in the early 1970s to mean a ``cassette tape'' and not only an ``adhesive tape''.
The change in the meaning of \texttt{tape} commences  with the introduction of magnetic tapes in 1950s (Figure \ref{fig:embeddings}).
The meaning continues to shift with the mass production of cassettes in Europe and North America for pre-recorded music industry in mid 1960s until it is deemed statistically significant.

The word \texttt{plastic} is yet another example, where the introduction of new products inflected a shift the word meaning. 
The introduction of Polystyrene in 1950 popularized the term ``plastic'' as a synthetic polymer, which was once used only denote the physical property of ``flexibility''.
The popularity of books on dieting started with the best selling book \emph{Dr. Atkins' Diet Revolution} by Robert C. Atkins in 1972 \cite{diet}.
This changed the use of the word \texttt{diet} to mean a life-style of food consumption behavior and not only the food consumed by an individual or group.

The \syn\ section of Table \ref{tab:ngrams_table} shows that words like \texttt{hug} and \texttt{sink} were previously used mainly as verbs.
Over time organizations and movements started using \texttt{hug} as a noun which dominated over its previous sense.
On the other hand, the words \texttt{click} and \texttt{handle}, originally nouns, started being used as verbs.

Another clear trend is the use of common words as proper nouns.
For example, with the rise of the computer industry, the word \texttt{apple} acquired the sense of the tech company Apple in mid 1980s and the word \texttt{windows} shifted its meaning to the operating system developed by Microsoft in early 1990s.
Additionally, we detect the word \texttt{bush} became widely used as proper noun in 1989, which coincides with George H. W. Bush's presidency in USA.
  
\subsection{Cross Domain Analysis}
Semantic shift can occur much faster on the web, where words can acquire new meanings within weeks, or even days. 
In this section we turn our attention to analyzing linguistic shift on Amazon Reviews and Twitter (content that spans a much shorter time scale as compared to \ngrams).

Table \ref{tab:internet_data} shows our \dist\ method results on Amazon Reviews and Twitter datasets.
New technologies and products introduced new meanings to words like \texttt{streaming, ray, rays, combo}.
The word \texttt{twilight} acquired new sense in 2009 concurrent with the release of the Twilight movie in November 2008.

Similar trends can be observed in Twitter.
The introduction of new games and cellphone applications changed the meaning of the words \texttt{candy}, \texttt{mystery} and \texttt{rally}.
The word \texttt{sandy} acquired a new sense in September 2012 weeks before Hurricane Sandy hitting the East Coast of USA.
Similarly we see that the word \texttt{shades} shifted its meaning with the release of the bestselling book \emph{``Fifty Shades of Grey''} in June 2012.

These examples illustrate the capability of our method to detect the introduction of new products, movies and books.
This could help semantically aware web applications to understand user intentions and requests better.
Detecting the semantic shift of a word would trigger such applications to apply a focused disambiguation analysis on the sense intended by the user.

\subsection{Quantitative Evaluation}
\label{sec:synthetic}
To evaluate the quantitative merits of our approach, we use a synthetic setup which enables us to model linguistic shift in a controlled fashion by artificially introducing changes to a corpus.

\begin{figure}[tb]
\centering
\begin{subfigure}{0.475\columnwidth}
  \includegraphics[width=\columnwidth]{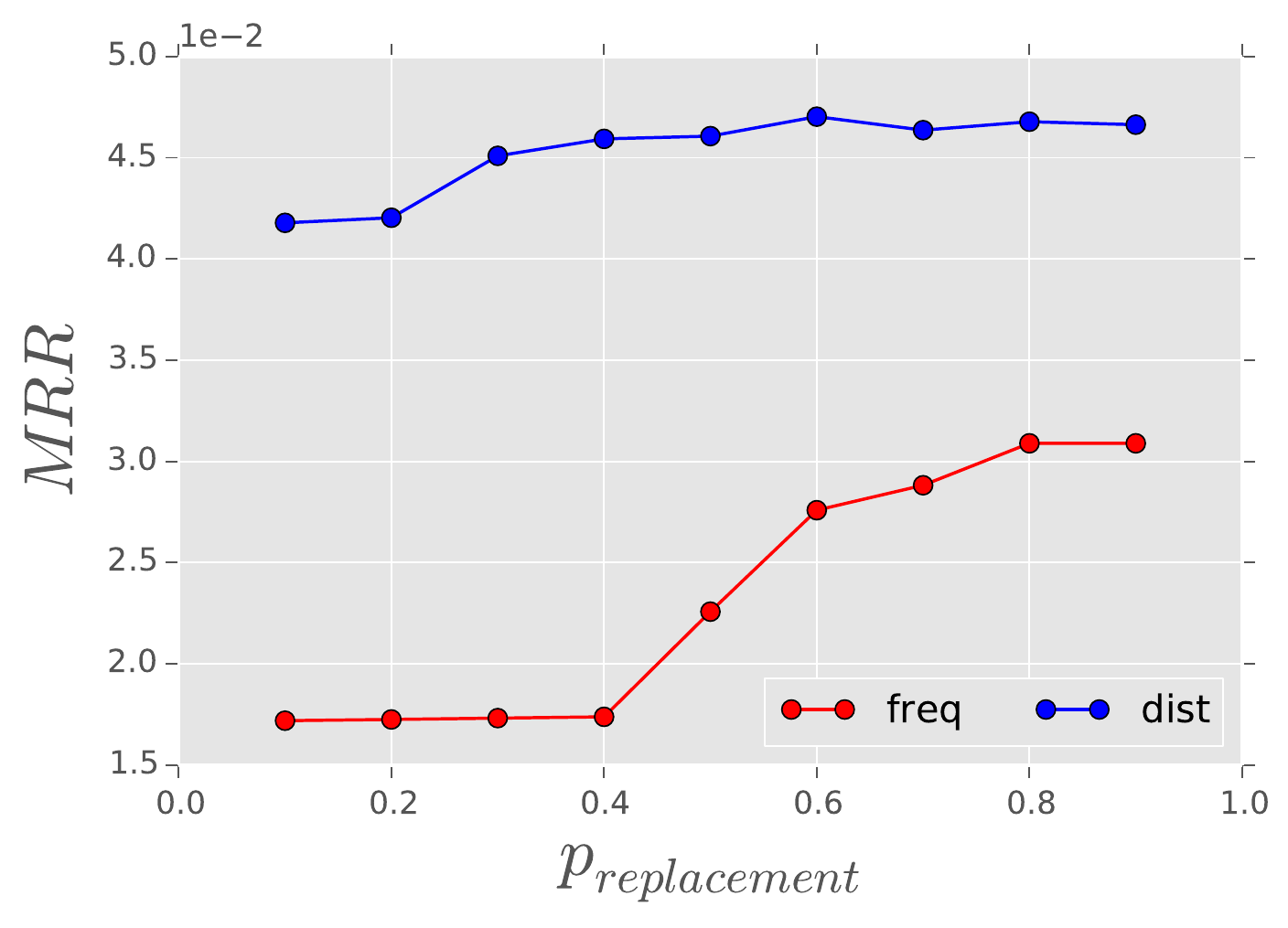}
  \caption{Frequency Perturbation}
  \label{fig:synthetic_freq}
\end{subfigure}
\begin{subfigure}{0.475\columnwidth}
   \includegraphics[width=\columnwidth]{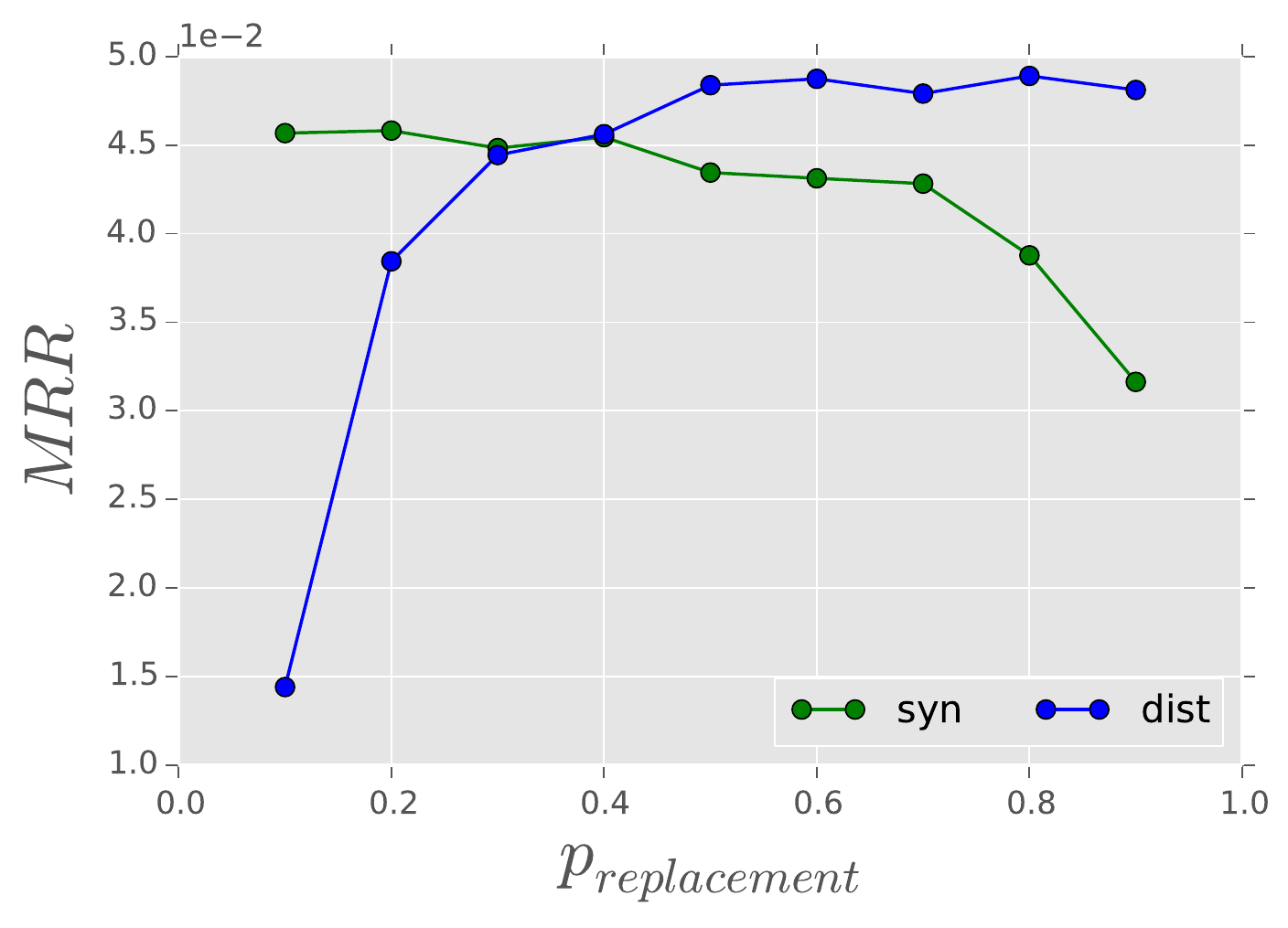}
   \caption{Syntactic Perturbation}
   \label{fig:synthetic_pos}
\end{subfigure}
\caption{Performance of our proposed methods under different scenarios of perturbation.}
\label{fig:synthetic}
\end{figure}

Our synthetic corpus is created as follows:
First, we duplicate a copy of a Wikipedia
corpus\footnote{\url{http://mattmahoney.net/dc/text8.zip}} 20 times to model time snapshots.
We tagged the Wikipedia corpora with part of speech tags using the
\emph{TextBlob} tagger\footnote{\url{http://textblob.readthedocs.org/en/dev/}}.
Next, we introduce changes to a word's usage to model linguistics shift.
To do this, we perturb the last 10 snapshots.
Finally, we use our approach to rank all words according to their $p$-values, and then we calculate the Mean Reciprocal Rank ($MRR = 1/|Q| \sum_{i=1}^{|Q|} 1/rank(w_i)$) for the words we perturbed.
We rank the words that have lower \pvalue\ higher, therefore, we expect the MRR to be higher in the methods that are able to discover more words that have changed.

To introduce a single perturbation, we sample a pair of words out of the vocabulary excluding functional words and stop words\footnote{NLTK Stopword List: \url{http://www.nltk.org/}}.
We designate one of them to be a donor and the other to be a receptor.
The donor word occurrences will be replaced with the receptor word with a success probability $p_{\text{replacement}}$.
For example, given the word pair (\texttt{location}, \texttt{equation}), some of the occurrences of the word \texttt{location} (Donor) were replaced with the word \texttt{equation} (Receptor) in the second five snapshots of Wikipedia.

Figure \ref{fig:synthetic} illustrates the results on two types of perturbations we synthesized.
First, we picked our (Donor, Receptor) pairs such that both of them have the same most frequent part of speech tag.
For example, we might use the pair (boat, car) but not (boat, running).
We expect the frequency of the receptor to change and its context distribution but no significant syntactic changes.
Figure \ref{fig:synthetic_freq} shows the MRR of the receptor words on \dist\ and \freq\ methods.
We observe that both methods improve their rankings as the degree of induced change increases (measured, here, by $p_{\text{replacement}}$).
Second, we observe that the \dist\ approach outperforms \freq\ method consistently for different values of $p_{\text{replacement}}$.

Second, to compare \dist\ and \syn\ methods we sample word pairs without the constraint of being from the same part of speech categories.
Figure \ref{fig:synthetic_pos} shows that the \syn\ method while outperforming \dist\ method when the perturbation statistically is minimal, its ranking continue to decline in quality as the perturbation increases.
This could be explained by the fact that the quality of the tagger annotations decreases as the corpus at inference time diverges from the training corpus.

It is quite clear from both experiments, that the \dist\ method outperforms other methods when $p_{replacement} > 0.4$ without requiring any language specific resources or annotators.

\section{Related Work}
\label{sec:related}
Because our work lies at the intersection of different fields, we will discuss the most relevant four areas of work:
linguistic shift, word embeddings, change point detection, and Internet linguistics.

\subsection{Linguistic Shift}
There has been a surge in the work about language evolution over time \cite{Michel:ngramscorpus,gulordava-baroni:2011:GEMS,cook2013,Wijaya:2011,heyer09,juola03languageChange}.
\citet{Michel:ngramscorpus} detected important political events by analyzing frequent patterns.
\citet{juola03languageChange} compared language from different time periods and quantified the change.
Different from both studies, we quantify linguistic change by tracking individual shifts in words meaning.
This fine grain detection and tracking still allows us to quantify the change in natural language as a whole, while still being able to interpret these changes.

Previous work on topic modeling and distributional semantics \cite{cook2013,Wijaya:2011,heyer09,gulordava-baroni:2011:GEMS,entropy} either restrict their period to two language snapshots , or do not suggest a change point detection algorithm. Some of the above work is also restricted to detecting changes in entities (e.g.\ Iraq). \citet{mitra-EtAl:2014:P14-1} use a graph based approach relying on dependency parsing of sentences. Our proposed time series construction methods require minimal linguistic knowledge and resources enabling the application of our approach to all languages and domains equally.
Compared to the sequential training procedure proposed by \citet{kim-EtAl:2014:W14-25} work, our technique warps the embeddings spaces of the different time snapshots after the training, allowing for efficient training that could be parallelized for large corpora.

Moreover, our work is unique in the fact that our datasets span different time scales, cover larger user interactions and represent a better sample of the web.


\subsection{Word Embeddings}
\citet{hinton1986learning} proposed distributed representations (word embeddings), to learn a mapping of symbolic data to continuous space.
\citet{bengiolm} used these word embeddings to develop a neural language model that outperforms traditional ngram models.
Several efforts have been proposed to scale and speed up the computation of such big networks \cite{hsm1,bengio2008adaptive,hsm2,dean:largescale}.
Word embeddings are shown to capture fine grain structures and regularities in the data \cite{mikolov2013linguistic,phrases}.
Moreover, they proved to be useful for a wide range of natural language processing tasks \cite{senna,polyglot}.
The same technique of learning word embeddings has been applied recently to learning graph representation \cite{deepwalk}.

\subsection{Change point detection}
Change Point Detection and Analysis is an important problem in the area of Time Series Analysis and Modeling.
\citet{Taylor00change-pointanalysis:} describes control charts and \textsc{CUSUM} based methods in detail.
\citet{adams-mackay-2007a} describes a Bayesian approach to Online Change Point Detection.The method of bootstrapping and establishing statistical significance is outlined in \cite{efron:bootstrap}.
\citet{Basseville:1993:DAC:151741} provides an excellent survey on several elementary change point detection techniques and time series models.

\subsection{Relation to Internet Linguistics}
Internet Linguistics is concerned with the study of language in media influenced by the Internet (online forums, blogs, online social media) and also other related forms of electronic media like Text Messaging.
\citet{teenuse} and \citet{ruin} study how teenagers use messaging media focusing on their usage patterns and the resulting implications on design of e-mail and Instant messaging (IM).
\citet{teenagers} study the language use by teenagers in online chat forums.
An excellent survey on Internet Linguistics is provided by \cite{Crystal:2011:ILS:2011923} and includes linguistic analyses of social media like Twitter, Facebook or Google+. 
\section{Conclusions And Future Work}
We proposed three approaches to model word evolution across time through different time series construction methods.
We designed a computational approach to detect statistically significant linguistic shifts.
Finally, we demonstrated our method on three different data sets each representing a different medium.
By analyzing the \ngrams, we were able to detect historical semantic shifts that happened to words like \texttt{gay} and \texttt{bitch}.
Moreover, in faster evolving medium like Tweets and Amazon Reviews, we were able to detect recent events like storms and game and book releases.
This capability of detecting meaning shift, should help decipher the ambiguity of dynamical systems like natural languages.
We believe our work has implications to the fields of Semantic Search and the recently burgeoning field of Internet Linguistics.

For our future work, we will focus on building real time analysis of data that link other attributes of data like geographical locations and content source to understand better the mechanism of meaning change and influential players in such change.
\label{sec:Conclusions}

\section*{Acknowledgments}
We thank Andrew Schwartz for providing us access to the Twitter stream data and for insightful discussions. 
\bibliographystyle{abbrvnat}
\bibliography{paper}
\end{document}